%% file: root.tex
\documentclass[journal, lettersize]{IEEEtran}  
\usepackage[switch]{lineno}
\usepackage{amsmath,amsfonts}
\usepackage{algorithm}
\usepackage{array}
\usepackage[caption=false,font=normalsize,labelfont=sf,textfont=sf]{subfig}
\usepackage{textcomp}
\usepackage{stfloats}
\usepackage{url}
\usepackage{verbatim}
\usepackage{graphicx}
\usepackage{cite}
\usepackage{tikz}
\usetikzlibrary{matrix,decorations.pathreplacing}
\IEEEoverridecommandlockouts

\usepackage{times} 
\usepackage{float}
\usepackage{caption}
\usepackage{tabularx,ragged2e,booktabs}
\usepackage{url}
\usepackage{todonotes}
\usepackage{algpseudocode}
\usepackage{graphicx}
\usepackage[font=footnotesize]{caption}
\DeclareMathOperator*{\argmax}{arg\,max}

\title{\LARGE \bf
Predictive Visuo-Tactile Interactive Perception Framework \\for Object Properties Inference
}
\author{Anirvan Dutta, Etienne Burdet and Mohsen Kaboli
\thanks{A Dutta and M Kaboli are with the BMW Group, RoboTac Lab, Munich, Germany. 
e-mail: name.surname@bmwgroup.com, https://www.robotact.de/}%
\thanks{A Dutta and E Burdet are with Imperial College of Science, Technology and Medicine, London, UK. M Kaboli is with Eindhoven University of Technology, Netherlands.}
\thanks{This work was supported by BMW Group, EU H2020
INTUITIVE under Grant ID 861166, and in part by EU Horizon PHASTRAC under Grant ID 101092096.}
}

\begin{document}
\bstctlcite{IEEEexample:BSTcontrol}
\maketitle
\thispagestyle{empty}
\pagestyle{empty}



\begin{abstract}
Interactive exploration of the unknown physical properties of objects such as stiffness, mass, center of mass, friction coefficient, and shape is crucial for autonomous robotic systems operating continuously in unstructured environments. Precise identification of these properties is essential to manipulate objects in a stable and controlled way, and is also required to anticipate the outcomes of (prehensile or non-prehensile)  manipulation actions such as pushing, pulling, lifting, etc. Our study focuses on autonomously inferring the physical properties of a diverse set of various homogeneous, heterogeneous, and articulated objects utilizing a robotic system equipped with vision and tactile sensors. We propose a novel predictive perception framework for identifying object properties of the diverse objects by leveraging versatile exploratory actions: non-prehensile pushing and prehensile pulling. As part of the framework, we propose a novel active shape perception to seamlessly initiate exploration. Our innovative dual differentiable filtering with Graph Neural Networks learns the object-robot interaction and performs consistent inference of indirectly observable time-invariant object properties. In addition, we formulate a $N$-step information gain approach to actively select the most informative actions for efficient learning and inference. Extensive real-robot experiments with planar objects show that our predictive perception framework results in better performance than the state-of-the-art baseline, and demonstrate our framework in three major applications for i) object tracking, ii) goal-driven task, and iii) change in environment detection.

\end{abstract}

\begin{IEEEkeywords}
Visual and Tactile Sensing, Active Interactive Perception, Recursive Bayesian Filtering
\end{IEEEkeywords}

\input{sections/introduction}

\input{sections/methods}

\input{sections/experiments}

\input{sections/conclusion}

\section*{Acknowledgment}
We would like to thank Dr. Xiaoxiao Cheng for his constructive reviews and feedback. In addition we thank Aitana Arranz Ibanez for helping to make 3D printed objects.

\bibliography{biblio}
\bibliographystyle{IEEEtran}
\input{sections/appendix}

\end{document}

%% file: sections/introduction.tex
\section{Introduction}
\label{sec:introduction}
\begin{figure}[t!]
    \centering
    \includegraphics[width=\columnwidth]{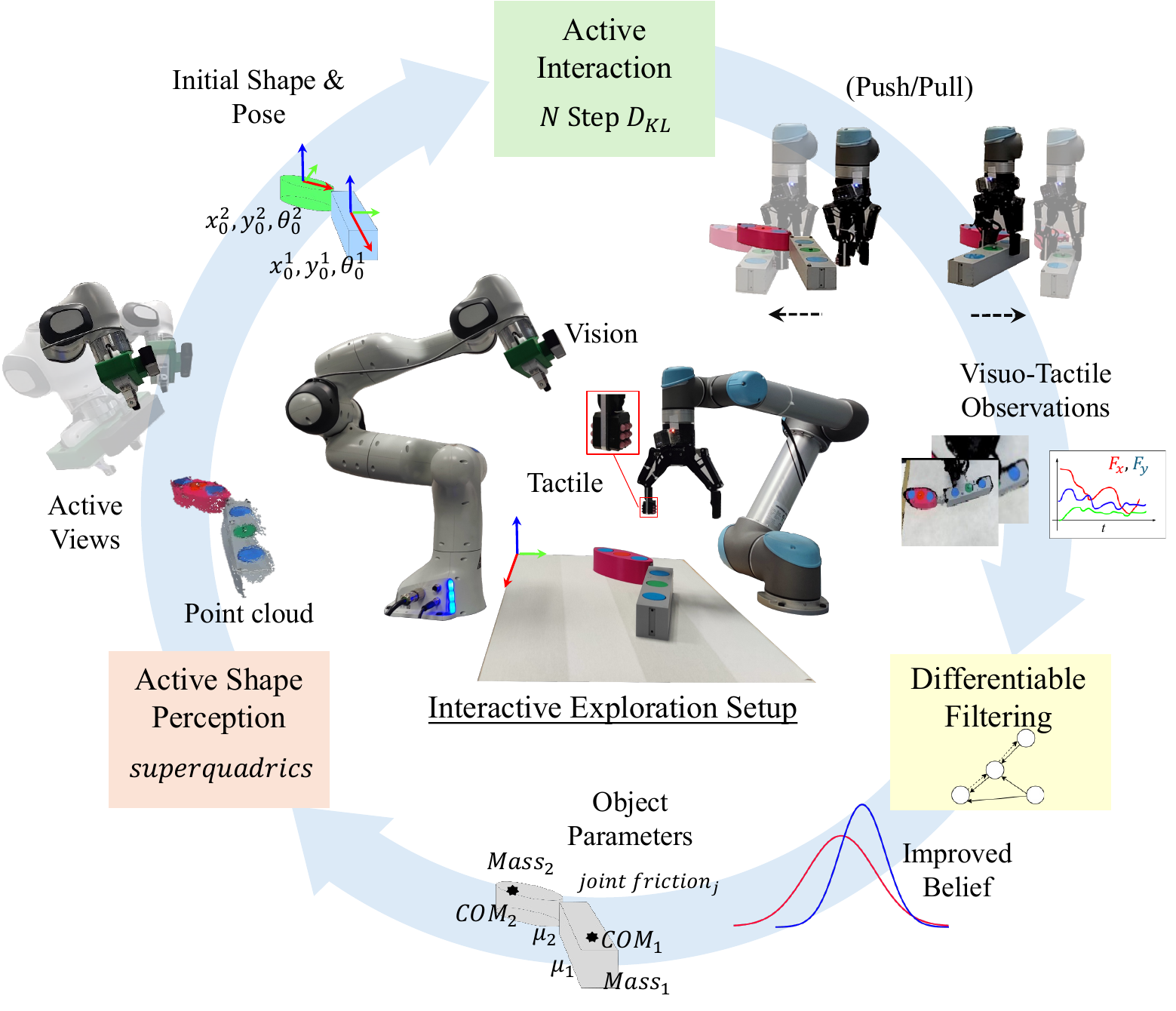}
    \caption{Overview of the proposed framework for visuo-tactile based interactive perception framework for active object exploration with three main components. 1. Uses visual information to actively estimate the shape of diverse objects based on superquadrics. 2. Actively selects the most informative action affordance for interaction. 3. Utilizes dual differentiable filtering for the estimation of objects' properties using visual and tactile information.}
    \label{fig:problem_setup}
\end{figure}

To increase the autonomy of the robotic system involved in various object manipulation tasks, it is essential that the robot perceive the physical properties of the object. However, estimating properties such as mass, the center of mass, and surface friction is challenging, as they are not directly observable in static environments and are salient only under specific object-robot interactions \cite{dense_phys}. Exploring previously unseen objects poses a challenge for current visual or tactile perception frameworks and necessitates the use of simple and robust interaction strategies. In this study, we introduce a novel predictive perception framework for inferring the properties of objects of various rigid objects such as with \textit{homogeneous}, \textit{heterogeneous}, \textit{articulated} properties and using vision and tactile sensing.

Previously, researchers have relied predominantly on vision or tactile methods to estimate the physical properties of objects. Tactile sensing offers a rich and diverse set of information about the object and allows the inference of multiple object properties. However, it requires precise information and prior knowledge ~\cite{kaboli_review_tro, seminara_active}. On the other hand, the range of properties observable through vision is limited ~\cite{navarro2023visuo}. Nevertheless, it can provide a global overview of the shape, and movement, and guide autonomous exploration. Recent works by Murali et al.~\cite{murali_interactive} and Lee et al. ~\cite{vt_lee_2021} have shown how a visuo-tactile-based approach can significantly improve the performance of robotic systems by addressing challenges such as pose estimation and contact-rich manipulation tasks. We aim to integrate complementary vision and tactile sensing to improve the reliability of robotic systems.

In such interactive visuo-tactile perception, purposeful physical interactions or explorations are made to improve object perception \cite{bohg_17_interactive, bajcsy_active, seminara_active,liu2022neuro,SharedVisuoTactile,ICRA-Mohsen}. Taking inspiration from an infant playful exploration \cite{agrawal2016learning,kaboli2016tactile,kaboli2017tactile, kaboli2018active} of pulling and pushing objects on the floor, our study focuses on two types of simple and natural exploratory actions: \textit{ non-prehensile pushing} and \textit{prehensile pulling} to explore the diverse set of objects. Pushing an object for a robotic system is a more straightforward task, particularly when dealing with large and heavy objects or when there is no prior knowledge about the object. In contrast, form-enclosure grasping \cite{danica_grasp} (prehensile) and pulling an object is a more stable approach compared to lifting and manipulating it, as factors such as the object's geometry and grasp stability come into consideration ~\cite{kaboli2016tactile,NeuromorphicScience,RAL-Mohsen}. Furthermore, for more sophisticated exploratory actions, complex robotic hardware would be required along with an intricate control mechanism. 

Nevertheless, exploring the properties of objects through non-prehensile push or prehensile pull poses a difficult challenge. This is due to the complexity of the dynamics of interaction between the object and the robot. Furthermore, the parameters are interrelated and there are significant uncertainties in both the contacts and the surface irregularities. To address such a challenging problem, we draw inspiration from neuroscience research, where such exploration behavior is inherent in humans \cite{barlow1990conditions}. A key working principle of human perception acknowledged since Helmholtz is that individuals actively predict perceptions relying on an internal model of their environment. Based on this prediction and immediate observation, humans make inferences online about their environment with associated uncertainties \cite{fep}.

In this study, we designed such a \textit{predictive} perception framework for a robotic system to learn and infer the properties of objects through interactive actions. The key aspect of our approach consists of encoding the object-robot interactions as Probabilistic Markov Models and learning the interaction model capable of predicting the visuo-tactile observation. This, compared with actual observation, will be used to estimate physical parameters such as mass, center of mass, and relative friction, since the physical properties of the objects cannot be observed directly. Bayesian filtering techniques coupled with the learnable model (differentiable filter \cite{oli_df_3}) are used as the foundation of the \textit{predictive} perception framework. 

We generalize and improve our previously proposed dual differentiable filter \cite{dutta2023push} that can be used to infer the time-varying object motion and time-invariant parameters consistently with pertinent uncertainty for diverse rigid objects with \textit{homogeneous}, \textit{heterogeneous}, \textit{articulated} properties. Furthermore, we leverage learnable noise models with the differentiable filter to detect changes in the learned model, which is not sufficiently researched in such interactive exploration setting. A critical aspect of such differentiable filters is the learned models. Given the diverse range of objects targeted, particularly with the inclusion of articulated objects, adopting a robust inductive bias inspired by physics becomes imperative in model selection. In this work, we leverage graph neural networks (GNNs) for this purpose, to encapsulate the intricate dynamics of interactions and elucidate how shape, pose, and physical properties contribute to visual and tactile observations. We propose a novel graph representation and propagation of graphs to model the interaction between a robot and an object. This representation focuses on the causal relationship of the interaction forces as the central element. It captures how the robot acts on the object (cause) while accounting for the environment in this interaction. This choice proves advantageous both in generalizing the model to be used in all the different cases of object-robot interaction presented in this work and also in capturing complex interactions sufficiently to account for tactile observations accurately (effect of object's movement and properties on the robot).

Furthermore, to improve data efficiency and inference time, it is essential for robotic systems to actively explore by strategically selecting the next-best exploratory actions. Previous works have shown that active object exploration outperforms a uniform and random strategy to reduce the uncertainty about objects while tackling different problems such as object recognition ~\cite{kaboli2019auro} and pose estimation ~\cite{vt_praj_2020}. In this study, we further evaluated our previously proposed non-greedy $N$-step Information Gain formulation \cite{dutta2023push} for active exploratory action selection and provide a more comprehensive analysis of this metric in various experimental scenarios, validating its effectiveness.

In addition, to ensure robust and seamless object exploration, we utilize \textit{superquadrics} for explicit shape representation and estimation, employing a Bayesian inference scheme \cite{liu2022robust}. This approach is advantageous because it requires no prior knowledge of shapes or primitives and effectively handles significant visual noise. Furthermore, compared to our previous approach in \cite{dutta2023push}, this approach effectively handles all the different object types. We prefer a low-dimensional shape representation \cite{wu2022primitive}, which is particularly beneficial for exploring novel objects \cite{palleschi2023grasp}, over traditional methods that produce high-dimensional point clouds or meshes necessitating complex post-processing. Additionally, we introduce a novel viewpoint selection method to improve the efficiency of shape estimation in a real robotic scenario where noisy and partial views of objects are unavoidable. The overview of our proposed interactive perception framework for active object exploration and inference of physical properties is illustrated in Fig. \ref{fig:problem_setup}.

\footnotetext{For supplementary materials, please visit \\
\textcolor{blue}{https://www.robotact.de/predictive-vistac}}

\section{Related Works}
\label{sec:relatedwork}
Estimating the physical properties of novel objects is a challenge in robotics, using either vision or tactile sensing, and the complexity is further increased when the object must be explored autonomously by a robotic system.

As a first step of object exploration, perception of its shape and pose is crucial for interaction. Object shape estimation or reconstruction has been extensively investigated in robotics, computer vision, and computer graphics employing statistical point-based, graph-based, view-based, and machine learning techniques. \cite{phang2021review}. Early vision-based shape completion aimed to fill missing areas in partial point clouds through local surface primitives \cite{schnabel2009completion, jenke2008surface}, or leveraging prior object structure and symmetry information \cite{bohg2011mind, thrun2005shape}. However, these approaches were often limited to specific categories of objects and lacked generalizability. To overcome these limitations, numerous machine learning-based shape completion methods have been developed \cite{fei2022comprehensive}. However, many of these methods rely on an offline collection of partial observations, often artificially created and not representative of real-world scenarios.

Few studies have explored multi-view or exploratory methods to enhance shape perception by determining the next-best view, a concept rooted in active vision \cite{aloimonos1988active}. In \cite{delmerico2018comparison}, various techniques for computing the next-best view, such as voxel and occupancy grid methods, were reviewed, highlighting Information Gain as a key factor in guiding active visual exploration. However, converting point clouds to volumetric data and employing ray casting is often slow, resource-intensive, and imprecise \cite{zeng2020view}. Recent advancements \cite{zeng2020pc, mendoza2020supervised, zhang2019reducing, jin2023neu} address these limitations by using learned models to align a prior with the current partial point cloud and estimate information gain for all candidate views, avoiding the need for extensive computations. Despite these improvements, these methods require substantial training and struggle with generalization, particularly for novel object exploration \cite{palleschi2023grasp}. Some studies \cite{potapova2020next, pineda2020active} have investigated reinforcement learning to generalize to unknown objects, but these approaches have primarily been limited to simulated environments. Consequently, a robust and effective active shape perception strategy for real-world robotic object exploration remains elusive.

Estimating the inertial and surface properties of rigid objects is a well-established problem in control theory, particularly in the context of identifying rigid body systems \cite{Ljung1998}, \cite{niebergall1997identification}. In these traditional settings, objects are typically constrained. In contrast, in robotics, requires autonomous exploration of objects in unstructured environments, presenting additional challenges not addressed by conventional identification methods. Earliest works in robotics such as \cite{atkeson_86, zhao_rel_inertia, nad_rel_inert} estimated the mass and moment of inertia of an object rigidly attached to a manipulator using joint torques or a wrist force torque sensor. Further studies have attempted to overcome the constraint of attaching the object to a manipulator using a specially designed mechanism with two fingers to measure contact forces during planar pushing ~\cite{yu_pushe_05} or a tilting approach to measure wrenches and estimate inertial parameters ~\cite{fukuda_pushe_99}. The study ~\cite{zhao_param_18} incorporated the estimation of friction by grasping the object and measuring the contact forces during the sliding regime. However, most of these previous estimation techniques relied on specialized mechanisms, required known object geometry, or incorporated assumptions regarding the interaction between the object and the environment. As a result, the generalization and autonomous exploration of objects became challenging.

Some researchers have tried to overcome the limitations mentioned above by introducing interactive manipulation techniques such as grasping or pushing. In \cite{tanaka_rel_04}, the authors estimated only the mass of an object by controlled push, which required prior knowledge of the friction coefficient of the surface. Similarly, the study \cite{kaboli_com_17} used tactile forces during a 3-finger robotic grasp to determine the center of mass of the object. To estimate the complete inertial matrix of a rigid object, the authors in \cite{bala_21} used a factor graph approach that involved in-hand manipulation with precise tactile sensing. This approach relied on approximations of the object's in-hand dynamics and knowledge of the object's shape and position. Previous studies that utilized interactive manipulation often employed an analytical approach to represent the interaction between the object and the robot. However, this approach is often based on approximations and relies on simplified assumptions about a specific robotic configuration.

In recent years, researchers have been exploring data-driven and physics-engine approaches to address these challenges. The authors of \cite{tenenbaum_data_15} used deep learning techniques to understand the interactions between objects colliding in a physics engine, where the learned model is utilized to estimate the mass and friction parameters of a real object. Similarly, \cite{song_19, song_20} used a physics engine to predict the expected motions of objects during pushing, and used Bayesian optimization on the actual motion of objects to estimate the distributed mass and friction on objects offline. However, these approaches heavily rely on the accuracy of the physics engine and are computationally complex. On the other hand, \cite{dense_phys} used vision and deep learning alone to learn a representation of the mass and friction coefficients by randomly pushing and poking objects. The study \cite{nikos_base_work} collected a large dataset of push trajectories (40k) in a simulation environment and trained a regression model to estimate an object's inertial parameters during non-prehensile pushing. The models learned in these works were limited to only homogeneous objects with uniform properties and could not be generalized to different types of objects.

Motivated by recent advances in graph neural networks (GNNs) \cite{sanchez2020learning, sanchez2018graph}, our goal is to leverage graphical models to learn the dynamics of interactions between objects and robots, enabling generalization across diverse object types. Studies such as \cite{tamim_graph, tekden2024object, wang2021dynamics} have shown the potential of graph networks in capturing complex object-robot interactions. For instance, \cite{tekden2024object} introduced a graphical model for estimating the pose and mass of objects in clutter through non-prehensile pushing. Current GNN methodologies primarily focus on spatial relationships among the objects, learning inherently the kinematics of the interaction. However, they fall short in capturing contact force details that depend not only depend on the spatial relation between the object and the robot, but it's physical properties, dynamics and also on the robot's actions. This highlights the need for a novel graph-based model that effectively handles these complexities to incorporate tactile information.

In addition, the data-driven methods mentioned above require extensive training and lack strategic interaction for model learning, which limits the use case in simulation environments. Although various recent works \cite{church2022tactile, jianu2022reducing} have addressed this by developing simulated tactile sensing environments with a low sim-to-real gap, they are often limited to vision-based tactile sensors or with static interaction.
The machine learning community has shown significant interest in active learning \cite{settles2009active}, particularly in strategies for effective labeling using an oracle or human annotator. Past research has predominantly focused on exploiting the uncertainty of classifiers or regressors to select the most effective unlabeled data, which are already present. In this study, we focus on active learning through efficient or informative actions, an area that is less explored in the literature \cite{taylor2021active}. In the studies \cite{kaboli2018active, kaboli2019auro}, the authors presented a Gaussian Process-based approach for learning and discriminating objects using active exploratory actions. Similarly \cite{xu2023tandem3d} trained an active exploration policy to discriminate a limited set of objects. More recently, the authors of \cite{sun2023active} have used uncertainty among the different classes of objects to actively explore and recognize objects. The above mentioned works are limited to classification setup with `greedy' one-step criterion's for action selection. Inspired by human `active' exploration and inference process \cite{parr2022active}, we introduced a novel $N$-step information gain formulation for both active exploration for learning and inference in \cite{dutta2023push}, which is evaluated for the diverse type and interactions of objects.

Until now, the estimation of physical object properties has mainly relied on vision or tactile sensing and offline estimation. Studies \cite{uttayopas2023object, le2021probabilistic, martin2022coupled} have demonstrated the effectiveness of online estimation using a probabilistic framework for robust and efficient parameter estimation, such as viscoelastic properties and friction, respectively. However, \cite{uttayopas2023object, le2021probabilistic} considered cases where the object was stationary, and in \cite{martin2022coupled}, the interaction with the object was teleoperated and the focus was solely on the tracking of articulated objects and the estimation of parameters. Moreover, none of the above approaches addressed the selection of strategic actions that take advantage of the probabilistic framework. 

In this study, we propose a novel visuo-tactile interactive framework to address the aforementioned challenges and limitations associated with estimating or inferring the physical properties of diverse types of objects. We actively perceive the shape and initial position of the object for seamless exploration. Thereby, we make use of non-prehensile pushing or prehensile pulling to explore the mechanical properties of the object. Furthermore, our proposed dual differentiable filtering with graph neural networks handles raw visuo-tactile observations in a predictive fashion and consistently performs inference on the position and parameters of the object. Our innovative $N$-step active formulation within the differentiable filtering framework enhances the efficiency of learning the object-robot interaction model and helps to select optimal exploratory push actions for efficient parameter estimation. We extend our previous work where we presented only non-prehensile pushing interaction for only homogeneous objects and improve our framework as follows

\subsection*{Our contributions are:}
I) We propose a novel active object shape perception approach leveraging Bayesian inference to estimate shapes as superquadrics for efficient exploration of object properties. The active method enables the robotic system to compute the next-best view for
the robotic system to look at the object to get complete shape information in challenging real-robotic exploration scenarios.

II) Our proposed differentiable filtering approach systematically addresses the time-invariant nature of object parameters and the time-varying object pose during exploration generalizing for both \textit{prehensile pulling} and \textit{non-prehensile pushing}.

III) We propose a novel graphical representation with a graph neural network (GNN)-based approach to capture intricate object-robot interaction models within the dual differentiable filter. This facilitates prediction of the visuo-tactile observation in advance and handles parameters of a diverse set of objects (such as homogeneous, heterogeneous, and articulated) with the same model.

IV) We further evaluate our innovative $N$-step look-ahead formulation exploiting the prediction step of the differentiable filtering for active action affordance selection for efficient learning the object-robot interaction model and inference of properties for different object types and interactions.  

V) We perform extensive real-robot experiments to validate the proposed method and compare it with the state-of-the-art baseline and present three key applications of the interactive perception framework: 1) Pose Estimation, 2) Goal-Driven Control, and 3) Change in Environment prediction.

%% file: sections/methods.tex
\section{Proposed Method}
\label{sec:method}
\begin{figure*}[ht!]
    \centering
    \includegraphics[width = \textwidth]{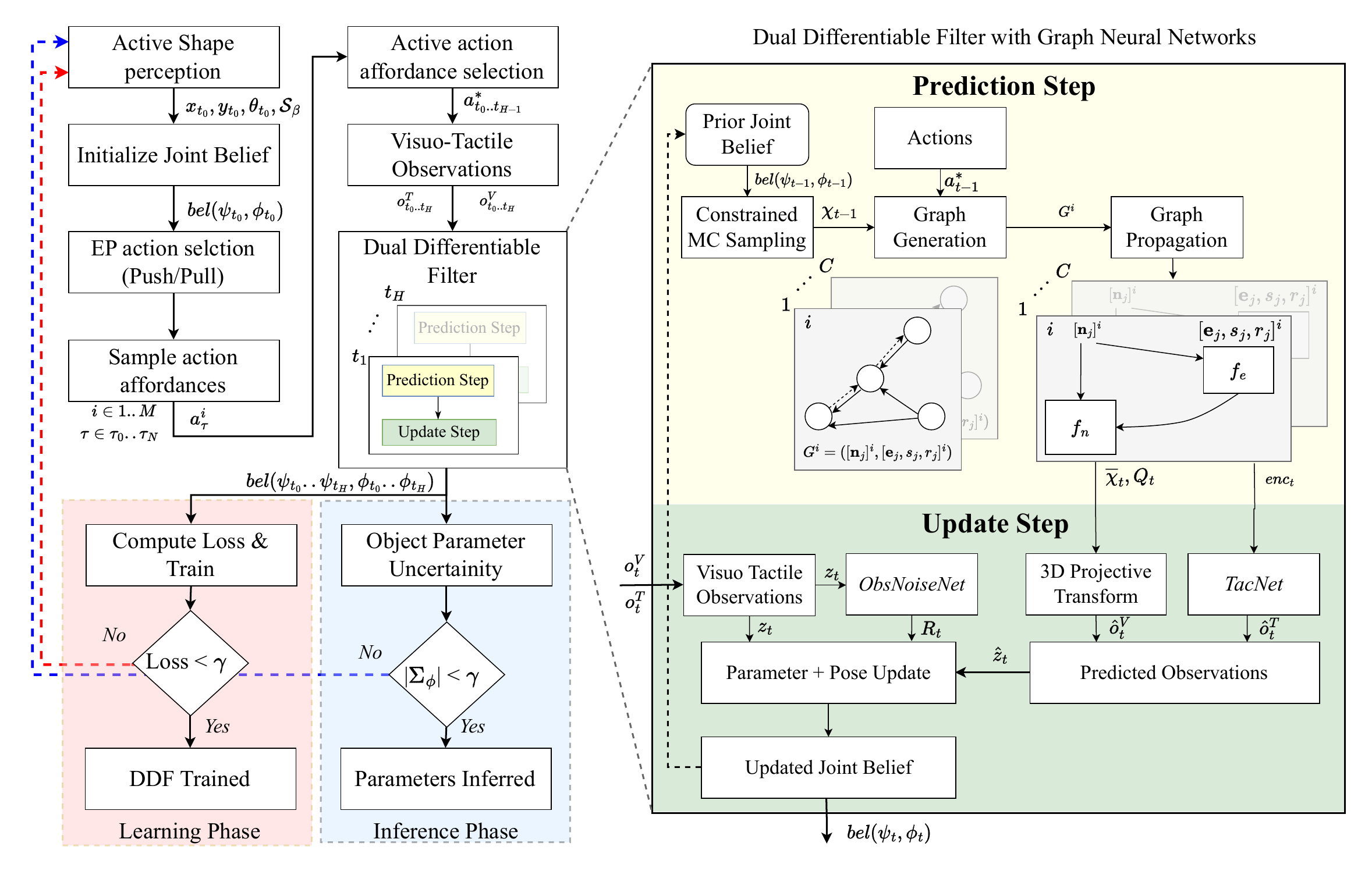}
    \caption{Our proposed framework is presented in detail for interactively inferring the diverse objects using visuo-tactile sensing. The framework starts in the learning phase followed by the inference phase.}
    \label{fig:framework}
\end{figure*}

\subsection{Problem Definition}
\label{subsec:prob_def}
We investigate the problem of estimating the state $s$ of an unknown rigid object placed on a support surface from visual $o^V$ and tactile $o^T$ observations using interactive actions $a$. The objects to be explored may consist of a single part or multiple parts (links), representing a wide range of objects found in our daily environment as presented in Fig. \ref{fig:experimental_object_set}. Therefore, an object can be a single link with uniform characteristics (\textit{homogeneous}), or it can have multiple links attached rigidly (\textit{heterogeneous}), or it can be an articulated object with multiple links connected by rotational joints (\textit{articulated}). 

To commonly represent uniform, heterogeneous, and articulated objects at any given time $t$, the state $s_t$ of the object is decomposed into $l \in {1, .. L}$ links $s_t = \{s^{1}_t ... s^{L}_t\}$. Each link $l$ state $s^{l}_{t} = \{ \psi^{l}_{t}, \phi^l \}$ consists of time-varying factors: 2D \textit{pose} and \textit{twist}, $\psi^{l}_{t}=\{x_t, y_t, \theta_t, v_{x_t}, v_{y_t}, \omega_t\}$ as well as time-invariant factors $\phi^l$, which is a combination of \textit{inertial parameters}: $\{m, CoM_x, CoM_y\}$ mass, vector of mass center w.r.t. the geometric center; as well as \textit{interaction parameters}: $\{f, f_r, f_{j} \}$ as friction-related parameters of with the support table, robot or subsequent link of the object, respectively. The 2D rotational inertia $I_z$ was found to not affect such quasistatic interactions and $f_r$ is assumed to be approximately known from the tactile sensor property. In addition, for autonomous and seamless exploration of the object, the shape of each link $\mathcal{S}^{l}$ is estimated via superquadrics. Although the primary focus is the estimation of the inertial and interaction parameters, we include the pose in the state, as the physical parameters are not directly observable and have to be estimated from the dependent pose. 

The observation $o^V_t$ comprises RGB-D images of the robot-object interaction area, and the tactile observation $o^T_t$ consists of \textit{ 2D contact forces} that are generated when the robotic gripper interacts with the object (fingertip forces). Two types of interactive exploratory actions on the basis of the shape and the initial pose of the object on the supporting surface. The pushing action is defined by the tuple \textit{contact point $(cp)$}, \textit{push direction} $(pd)$ and \textit{push velocity} $(u)$ of the push. $cp$ consists of the 2D world coordinate of the contact point, $pd$, the rotational angle of the z-axis of the robotic system aligned along a pushing direction \& $v$ is the magnitude of the push velocity by the robotic system. Similarly, the prehensile pull action is defined by a tuple of: \textit{ grasp point $(gp)$}, \textit{pull direction} $(pd)$ and \textit{pull velocity} $(u)$. 

The proposed framework is illustrated in Fig. \ref{fig:framework}.  It starts with active shape perception to identify the shape, links, and initial pose of the object. Thereby, one of the exploratory actions is selected with active affordances. This action is executed, and visuo-tactile observations are utilized within the dual-differentiable filtering to estimate the state of the object through the interaction. First, the robotic system learns the data-driven graphical model leveraged within the dual differentiable filter. After learning, inference on unknown objects is made to estimate the parameters. In the following sections, we explain the various components of the framework.

\subsection{Active Shape Perception}
\label{subsec:asp}
\subsubsection{Superquadrics}
\begin{figure}[h]
    \centering
    \includegraphics[width = \columnwidth]{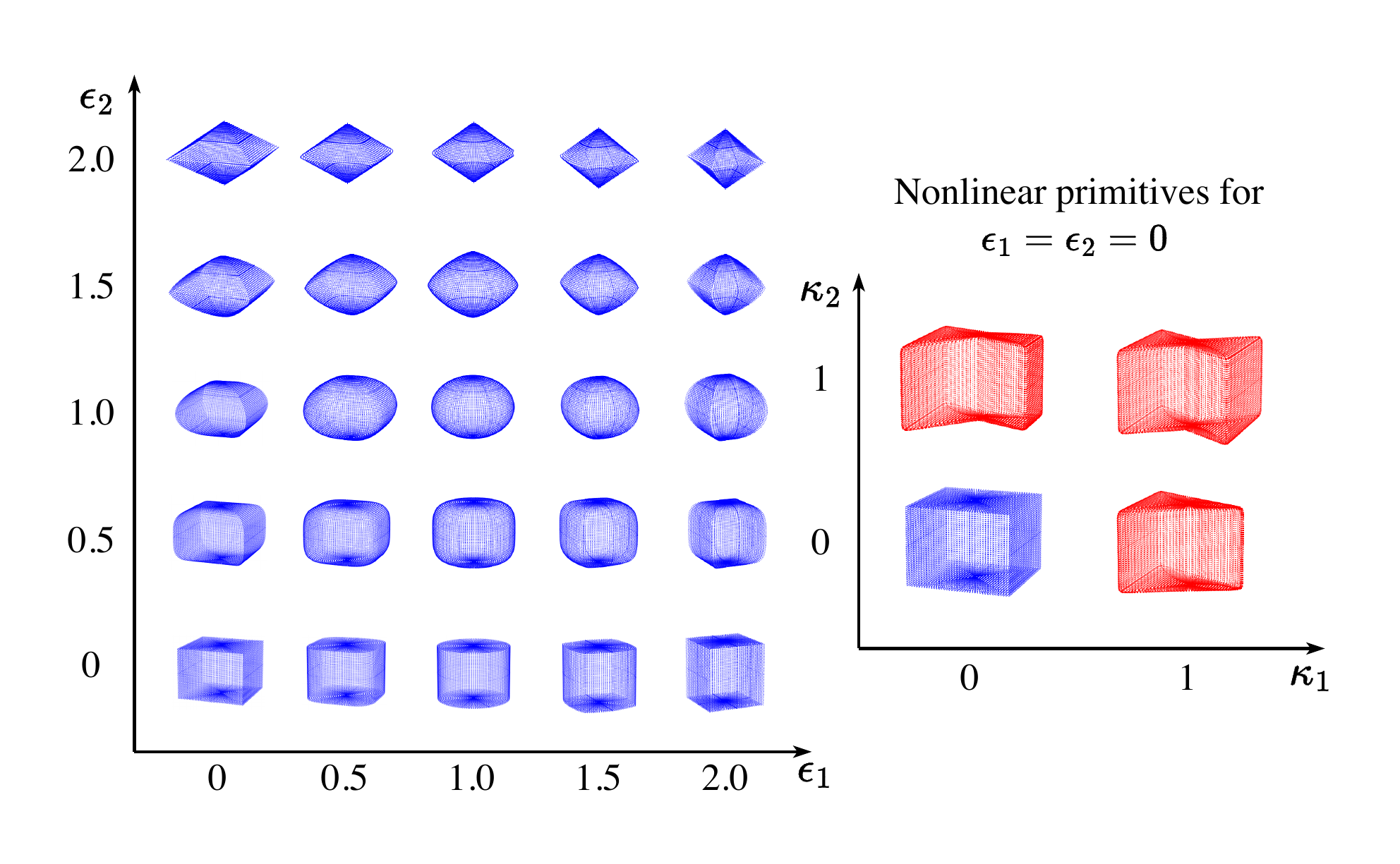}
    \caption{Illustration of a few basic superquadric shapes with the proposed non-linearity for one of the primary shapes}
    \label{fig:superquadric_basic}
\end{figure}

As a first step towards exploring the properties of an object, the shape of each link is estimated $\mathcal{S}^{l}$ using superquadrics, which are a family of geometric primitives. Superquadrics offer a rich shape vocabulary, such as cuboids, cylinders, ellipsoids, octahedra, and their intermediates, encoded by only five parameters. A superquadric centered in the origin with a frame aligned with the global x, y, z co-ordinate follows the following implicit function \cite{barr_sq}:
\begin{align}
    F(x, y, z) = {\left(\left(\frac{x}{a_x}\right)^{\frac{2}{\epsilon_2}} + \left(\frac{y}{a_y}\right)^{\frac{2}{\epsilon_2}}\right)}^{\frac{\epsilon_2} {\epsilon_1}} + \left(\frac{z}{a_z}\right)^{\frac{2}{\epsilon_1}}
\end{align}

\noindent where $\textbf{x} = [x, y, z]^T \in \mathbb{R}^3$ is a point or surface vector defined in the superquadric frame. Exponents ($\epsilon_1, \epsilon_2$) produce a variety of convex shapes and describe the shaping characteristics. As superquadrics are restricted to symmetric shapes only, a nonlinear deformation parameter to model non-convexity is proposed to represent a more general form of objects. A nonlinear tapering deformation is introduced along the y-axis of the superquadric frame as follows:

\begin{equation}
    f(y) = \kappa_1 y + \kappa_2 y^2, \quad X = f(y) x: Y = y: Z = z
    \label{eq:nonlinearshape}
\end{equation}

\noindent where $-1 \leq \kappa_1 \leq 1$ and $0 \leq \kappa_2 \leq 1$., $X, Y, Z$ are the components of the surface vector $\mathbf{X}$ of the deformed superquadric, $f(y)$ is the tapering function, $x, y, z$ are the components of the original surface vector $\mathbf{x}$. Fig. \ref{fig:superquadric_basic} illustrates the range of superquadrics, along with deformation. 

We can fully parameterize a superquadric placed on the table with parameters $\beta = [{\epsilon_1, \epsilon_2, a_x, a_y, a_z, \kappa_1, \kappa_2, g}]$ where $g=[x_0, y_0, \theta_0]$ is the initial 2D pose of the object. We aim to robustly optimize the superquadric parameters from the noisy and partial point cloud obtained from multiple visual observations $o^{V}$. For this, Expectation Maximization is utilized based on the work of \cite{liu2022robust}, which requires casting the estimation problem as a Bayesian inference problem. This also enables us to compute the uncertainty over the shape more accurately and perform novel active next-best-view computation to effectively improve the superquadric parameters.  




\subsubsection{Estimation of parameters of superquadric}
A Gaussian centroid $\mathbf{\mu} \in \mathcal{S}^{l}_{\beta}$, randomly sampled on the superquadric surface $\mathcal{S}^{l}_{\beta} \subset \mathbb{R}^3$ and parameterized by $\beta$, according to the uniform density function:
\begin{equation} 
p(\mathbf{\mu}) = \frac{1}{A_{\beta}}, \qquad A_\beta = \int\limits_{\mathcal{S}^{l}_\beta} 1d\mathcal{S}
\label{eq:sq_sample}
\end{equation} 
\noindent where $A_\beta$ is the area of superquadric. An observation $\mathbf{x}$ from a Gaussian-uniform model is created in $\mathbb{R}^3$ with a probability density function:

\begin{equation} 
p(\mathbf{x} | \mathbf{\mu}) = w_0p_0(\mathbf{x}) + (1-w_0)\mathcal{N}(\mathbf{x} | \mathbf{\mu}, \Sigma) 
\label{eq:shape_main}
\end{equation} 

\noindent where $\mathcal{N}(.| \mathbf{\mu}, \Sigma)$ denotes the density function of a Gaussian distribution parameterized by ($\mu, \Sigma$). Noise is assumed to be isotropic, i.e. $\Sigma = \sigma^2I$ and $w_0 \in [0, 1]$ represents the probability that a point is sampled from an outlier distribution - $p_0(\mathbf{x}) = 1/V$, with $V$ encapsulating the volume of the interaction area. Eq. \ref{eq:shape_main} is simplified by introducing a latent discrete random variable $\gamma$ that serves as an indicator of the membership of $\mathbf{x}$. When $\gamma=0$ $\mathbf{x}$ is sampled from the uniform outlier component, while, $\gamma=1$ indicates that $\mathbf{x}$ is generated from the Gaussian inlier component. This reformulation results in the following:

\begin{align}
    p(\mathbf{x} | \mathbf{\mu}, \gamma) &= p_0(\mathbf{x})^{1-\gamma}\mathcal{N}(\mathbf{x} | \mu, \Sigma)^\gamma \\
    \gamma \sim p(\gamma) &= Bernoulli(1-w_0)
\end{align}

Given a set of points from the point cloud $\mathbf{X} = {\mathbf{x}_i \in \mathbb{R}^3 |i = 1, 2, .. N}$, the parameters of the superquadric surface can be estimated by minimizing the negative log-likelihood function:

\begin{equation}
    l(\beta, \sigma^2) = \sum_{i=1}^{N} \gamma_i \left(\frac{||\mathbf{x}_i - \mu_i||{^2}_2}{2\sigma^2} -log c\right) + Nlog(A_\beta)
    \label{eq:shape_mle_problem}
\end{equation}

\noindent where $c$ is the normalization constant of the Gaussian distribution. The authors in \cite{liu2022robust} employed a novel Expectation Maximization coupled with a Switching (EMS) approach to solve the MLE problem and overcome the local optimality in Eq.\ref{eq:shape_mle_problem}. The switching approach generates similar candidate superquadrics which are then replaced when the EM algorithm gets stuck in the local minima. In our case, along with the general generation of superquadric candidates as in \cite{liu2022robust}, similar candidates due to nonlinear deformation terms were also generated. The EMS approach first estimates the Gaussian centroids given the current estimation of the superquadric parameters. 
\begin{equation}
    \hat{\mu}_i = \underset{\mu_i \in \mathcal{S}^{l}_{\beta}}{\mathrm{argmin}} 
 p(\mu_i|x_i) \sim  \underset{\mu_i \in \mathcal{S}^{l}_{\beta}}{\mathrm{argmin}} ||x_i - \mu_i||^2
 \label{eq:shape_mu}
\end{equation}



With the current estimate of $\mu_i$, the expectation of the posterior probability of $x_i$ being an inlier is inferred via the Bayes' rule: 

\begin{equation}
    E(\gamma_i=1|x_i, \hat{\mu}_i) = \frac{\mathcal{N}(x_i |\hat{\mu}_{i}, \sigma^2I)}{\mathcal{N}(x_i|\hat{\mu}_i, \sigma^2I) + \frac{w_op_o(x_i)}{1-w_o}}
\end{equation}

Finally, the parameters $\beta, \sigma$ are optimized by the M-Step by substituting the posterior estimates $\hat{\mu}_{i}$ and $\gamma_i$ in Eq. \ref{eq:shape_mle_problem}. The EM step is applied iteratively until the change in the parameters $\beta, \sigma$ is less than the threshold $(0.001)$. 

\begin{figure*}[t!]
    \centering
    \includegraphics[width = 0.85\textwidth]{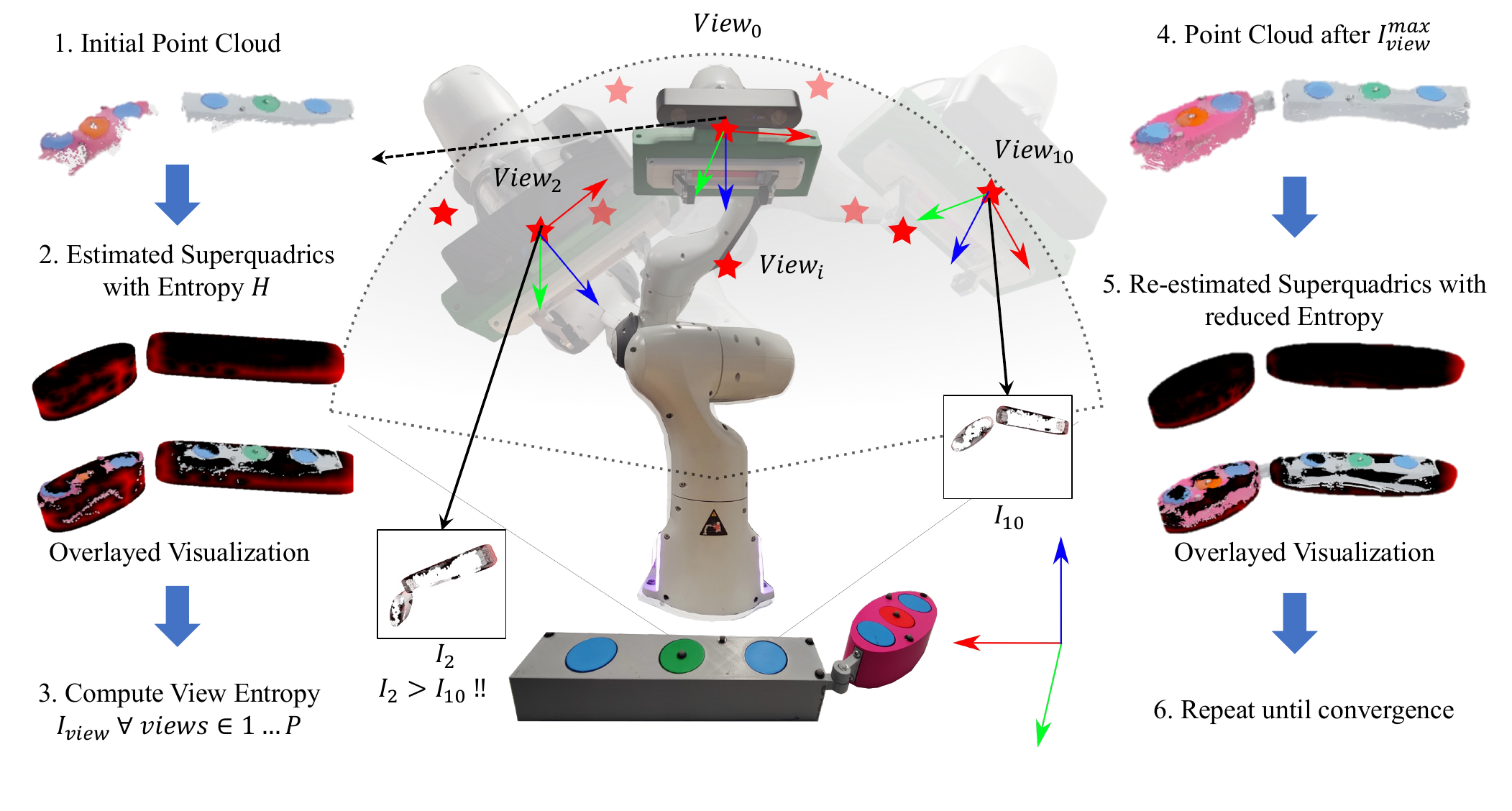}
    \caption{Illustration of the shape perception approach. The entropy of each point calculated from Eq. \ref{eq:fullentropy} is mapped to 0-255 red channel of the sampled superquadrics, a higher red indicates higher entropy. Viewpoint entropy computed from Eq. \ref{eq:projective_transform} of 2 sampled points $I_{2}$ and $I_{10}$ are also presented}
    \label{fig:activeshape}
\end{figure*}

\subsubsection{Multi-Superquadric Recovery}
This approach is extended to identify multiple links $L$ of an object from the initial point cloud $\mathbf{X}$. When the estimated superquadric fails to sufficiently fit parts of the point cloud, those points are classified as outliers $\gamma=0$. Clustering these outlier points allows for iterative estimation of new superquadrics, continuing until the point cloud's maximum coverage is achieved i.e. number of points in outlier a used defined threshold $< O_{th}$). The number of links $L$ identified is the total number of superquadrics inferred from the point clouds.  

\subsubsection{Next-Best View: Viewpoint Entropy}
Having discussed the Bayesian inference approach to estimate the parameters of superquadrics, we now present a novel approach to compute the next-best view.  In robotic scenarios, it is often the case that only a partial point cloud of the object is available from one view, as illustrated in Fig. \ref{fig:activeshape}. To improve the estimation of the superquadric parameters, it is necessary to compute multiple next-best views. We propose a novel active approach, which uses the current parameters of the superquadric as a prior to compute the expected entropy of a hypothetical view. Unlike other ray casting approaches \cite{vt_praj_2020}, this approach i) does not require any approximation of occupancy grids, ii) can compute the entropy more accurately as it considers the current shape, and iii) can account for camera intrinsic. 

We uniformly sample points $\mathbf{\mu}_j | j = 1,2,..M$ on the surface of the each estimated superquadrics $\mathcal{S}^{l}_{\beta} | l = 1, 2.. L$ using the approach presented in \cite{liu2022robust}. We then compute the expected entropy of the superquadric surface given the current partial point cloud points $\mathbf{x}_i | i = 1, 2.. N$:
\begin{equation}
    H(\mathbf{\mu}_j) = -p(\mu_j|\mathbf{x}_i)log(p(\mu_j|\mathbf{x}_i))
    \label{eq:fullentropy}
\end{equation}
$p(\mu_j|\mathbf{x}_i)$ is approximated as:
\begin{equation}
    p(\mu_j|\mathbf{x}_i) \sim \underset{\mathbf{x_i}}{\mathrm{argmin}} ||\mathbf{x} - \mu_j||^2
    \label{eq:approximation}
\end{equation}

\noindent which is the minimum distance between the sampled point and a point in the point cloud. We define a view point $a^{view}$ by a 3D position $\mathbf{p}^{view} \in \mathrm{R}^3$ and orientation $\mathbf{R}^{view} \in SO(3)$.  Subsequently, $P$ viewpoints are uniformly sampled around the workspace on the hemisphere space located about the object. The 3D position $\mathbf{p}^{view}$ is sampled as a point on the surface of the hemisphere and the orientation $\mathbf{R}^{view}$ is autonomously computed so that the camera is expected to `look at' the object. It is computed as the axis of rotation $\mathbf{\hat{e}}$ and the angle $\theta$ given by:
\begin{align}
   \mathbf{\hat{h}} = & \frac{\mathbf{p}^{view} - [\hat{x}_0, \hat{y}_0, 0]}{||\mathbf{p}^{view} - [\hat{x}_0, \hat{y}_0, 0]||} \\
     \theta = cos^{-1}(\mathbf{\hat{h}}. \mathbf{\hat{Z}}) &\qquad \mathbf{\hat{e}} = \frac{\mathbf{\hat{h}} \times \mathbf{\hat{Z}}}{||\mathbf{\hat{h}} \times \mathbf{\hat{Z}}||}
\end{align}

We use projective transformation \cite{mcglone2013manual, nematollahi2022t3vip} method to convert the entropy for each sampled point in 3D using Eq.\ref{eq:fullentropy} to a sampled camera $I^{view}$, which is referred as \textit{view entropy}:
\begin{align}
I^{view} = \sum_{j=1}^{M} K [\mathbf{R}^{view} \mathbf{p}^{view}] \begin{bmatrix}
H(\mathbf{\mu}_{x_j})\\
H(\mathbf{\mu}_{y_j}) \\
H(\mathbf{\mu}_{z_j}) \\
1
\end{bmatrix}
\label{eq:projective_transform}
\end{align}

where, $K$ is the intrinsic matrix and $[\mathbf{R}^{view} \mathbf{p}^{view}]$ is the homogeneous extrinsic camera matrix. The viewpoint $a^{view}$ with the maximum entropy value $I^{view}_{max}$ is selected as the next best view to obtain the next visual observation or the point cloud. Subsequent point clouds are registered using ICP \cite{icp_old} to overcome residual camera calibration errors, and superqadric parameters are recomputed.  
Active viewpoint selection and subsequent shape estimation are illustrated for an example object in Fig. \ref{fig:activeshape}. The process is terminated when the computed entropy of the superquadric is reduced below a threshold. We present both quantitative and quantitative results on shape perception in Section~\ref{sec:results} and in the Appendix. The recovered shape parameters are used to sample the locations of the action locations (contact points) for push and pull, and also initialize the joint belief with the initial pose information of each link. 

\subsection{Differentiable Filters}
\label{subsec:df}
After the shape of the object, we now present the filtering formulation for interactive exploration. We represent the belief about the current state of the object $s_t$ with a distribution conditioned on previous actions $a_{1:t}$ and observations $o_{1:t}$. This distribution is denoted as the belief of the state  
\begin{align}
\begin{split}
     bel(s_t) & = p(s_t|o_{1:t}, a_{1:t}) \\
     & = \frac{p(o_t|s_t,o_{1:t-1}, a_{1:t})p(s_t|o_{1:t-1}, a_{1:t})}{p(o_t|o_{1:t-1},a_{1:t})}
 \end{split}
 \label{eq:belief_state}
 \end{align}
One prominent approach to computing the belief tractably is to employ Recursive Bayesian Filters which use the Markov assumption i.e. that the future belief of the state is conditionally dependent only on the current state to simplify Eq.\ref{eq:belief_state}  This yields a recursive structure:
\begin{align}
\begin{split}
    bel(s_t) & = \eta p(o_t|s_t,a_t)\int_{}^{}p(s_t|s_{t-1}, a_{t-1})bel(s_{t-1})ds_{t-1} \\
    & = \eta p(o_t|s_t,a_t)\overline{bel}(s_t)   
\end{split}
\label{eq:bayes_filter}
\end{align}

\noindent where $\eta$ is a normalizing factor. Kalman Filters are a common choice of Bayesian Filtering which is optimal in linear systems and can be extended in nonlinear cases using various approaches \cite{prob_rob_book}. Two key aspects of Bayesian filtering are the representation of the state process model in the form of $p(s_t|s_{t-1}, a_{t-1})$ and an observation likelihood model that relates the states to the observations $p(o_t|s_t)$. For our problem, a data-driven approach is used to learn the process and the observation model along with the observation noise model. As the pose of the object is intricately dependent on the inertial and interaction parameters, straightforward combined (joint) filtering for pose and parameters does not perform well. Therefore, we utilize a dual filter design, exploiting the dependency among the states for consistent filtering and inferring the parameters of the object. We present the action selection approach followed by the dual filtering methodology.

\subsection{Exploratory Action Selection}
\label{subsec:activeaction}
Considering the constraints of the workspace, the robotic system selects the exploratory prehensile or non-prehensile interaction. When the initial pose of the object leans towards the edge of the table ($y_0 < 0.3$), prehensile pulling is selected. In contrast, if the object is placed ($y_0  \geq 0.3$) closer to the robot, the push action is selected. Additionally, we found certain homogeneous objects that were too small in shape, leading to substantial occlusion in the vision system, or challenging to perform prehensile grasp with a 2-finger gripper (such as the triangular object). Consequently, the robotic system exclusively uses non-prehensile pushing for homogeneous objects $L=1$, while using both prehensile pulling and non-prehensile pushing for heterogeneous and articulated objects $L>1$.

\subsection{Active Action Affordance Selection}
\label{subsec:activeaffordance}
The action affordance for the push action is $a_t = (cp, pd, u)$ and the pull action is given by the tuple $a_t = (gp, pd, u)$ as presented in the problem definition. The possible \textit{ contact point} $cp$ or grasp point $gp$ and the normal angle $cn$ at the contact point are calculated from the estimated superquadric $\mathcal{S}^l_{\beta}$. We present a combined active action affordance $a_t$ formulation of non-prehensile pushing and prehensile pushing as follows:

\textit{Monte-Carlo Sampling of action affordances}:
$M$ action affordances are generated, $a^{i}_t | i \in 1.. M$,  from the possible points of contact points and contact normal by sampling a contact point and generating the $pd^{i} = cn^{i} + \delta; \delta \sim R(-5, 5)$ (deg). The velocity $v$ is fixed for all cases, keeping in mind the quasi-static assumption. 

\textit{$N$-step Information Gain}
To make the framework more sample efficient for real robot scenarios, we leverage active action selection by formulating an $N$-step information gain criterion in the filtering setting. We recursively use the process model $p(s_t|s_{t-1}, a_{t-1})$ of the differentiable filter to compute the expected Information Gain for both model learning and object parameter inference for each sampled action $\pi^{i}$ = $a^i_{\tau_0:\tau_N}$ over the $N$-step in future $\tau = \tau_0 .. \tau_N$ 
\begin{align}
     \resizebox{.95\hsize}{!}{$IG_{N}(\pi^{i}) \approx  - \mathbb{E}_{p(\psi_{\tau_N}, \phi_{\tau_N}|\pi^{i})}[ln(\overline{bel}^{i}(\psi_{\tau_N}, \phi_{\tau_N}) - ln(\overline{bel}^{i}(\psi_{\tau_0} , \phi_{\tau_0})]$}
     \label{eq:IG_N_step}
\end{align}
\noindent where $\overline{bel}^{i}(\psi_{\tau_N}, \phi_{\tau_N})$ is the hypothetical predictive joint distribution after $N$-step by taking action $\pi^{i}$ without taking account the actual observation. For our case, the expectation is computed as KL-Divergence for which the closed form solution exists for Multivariate Gaussian distributions \cite{kl_gauss}. 

\begin{align}
    \resizebox{.95\hsize}{!}{$IG_{N}(\pi^{i}) \approx D_{KL}[\mathcal{N}^{i}(\psi_{\tau_N}, \phi_{\tau_N} | \overline{\mu}_{\tau_N}, \overline{\Sigma}_{\tau_N})
     || \mathcal{N}^{i}(\psi_{\tau_0} , \phi_{\tau_0} | \overline{\mu}_{\tau_0}, \overline{\Sigma}_{\tau_0})]$} \nonumber
     \label{eq:active_action}
\end{align}
\begin{equation}
    \pi^* = \argmax_{\pi^i} IG_N(\pi^{i})
\end{equation}

The details of the process model $p(s_t|s_{t-1}, a_{t-1})$ used to calculate the expected belief $\overline{bel}^{i}(\psi_{\tau_N}, \phi_{\tau_N})$ are presented in the following section. After the action is executed, the dual filtering step is followed.

\begin{figure*}[tb!]
    \centering
    \includegraphics[width = 0.85\textwidth]{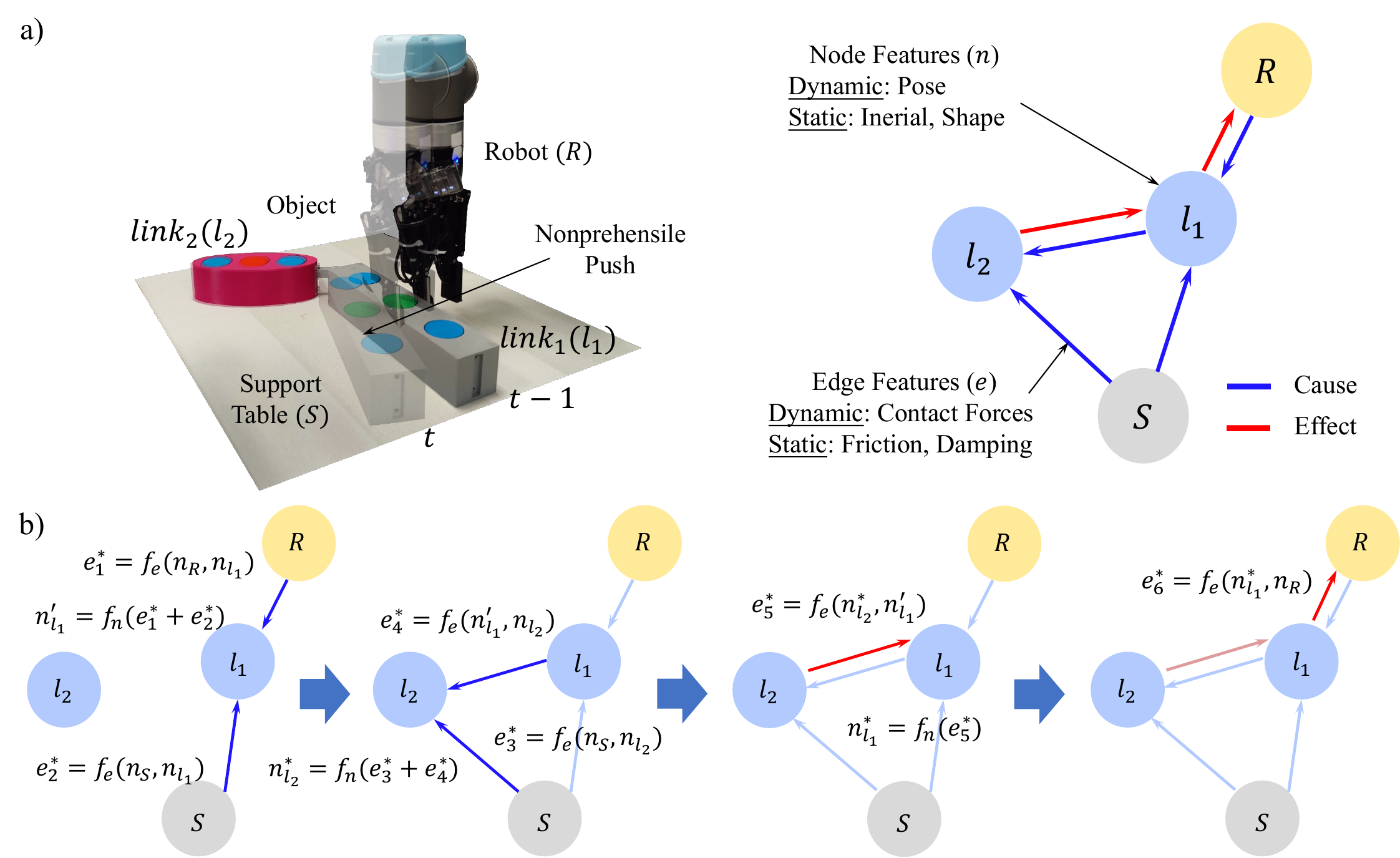}
    \caption{a) Illustration of the proposed graph representation of an example articulated object with two links b) Novel graph propagation for updating the graphical model from time $t-1$ to $t$ for the example object. The support edges $e_1, e_2$, the edge $e_6$ contains contact force or tactile information [Improve]}
    \label{fig:graphnets}
\end{figure*}

\subsection{Dual Differentiable Filter}
\label{subsec:ddf}
We derive our dual filter based on differentiable UKF \cite{prob_rob_book, alina_2_main}, which has been shown to perform better than EKF or particle filters for object tracking during tabletop manipulation \cite{alina_2_main}. For the dual filter formulation, we explicitly represent the state of the object with $s_t = \{s^{1}_t, ... s^{L}_t\}$ by the joint distribution of pose and twist $\psi_t = \{\psi^{1}_{t} .. \psi^{L}_{t}\}$ and combination of inertial and interaction parameters $\phi_t  = \{\phi^{1}_{t} .. \phi^{L}_{t}\} $ via Multivariate Gaussian distribution: 
\begin{equation}
     bel(\psi_t, \phi_t) \doteq \mathcal{N}(\psi_t, \phi_t | \mu_t , \Sigma_t)
     \label{eq:joint_distribution}
 \end{equation}
\begin{align}
    \mu_t = \begin{pmatrix}
\mu_{\psi_t} \\
\mu_{\phi_t}
\end{pmatrix}, \quad \Sigma_t = \begin{pmatrix}
\Sigma_{\psi_t} & \Sigma_{{\psi_t\phi_t}}\\
\Sigma_{{\phi_t\psi}_t} & \Sigma_{\phi_t}
\end{pmatrix} \quad .
\end{align}

Thee \textit{homogeneous}, \textit{heterogeneous} and \textit{articulated} can be commonly represented as objects with $L$ links, $\phi^{l}_t = \{x_t, y_t, \theta_t, v_{x_t}, v_{y_t}, \omega_{t}\}$ and $\psi^{l}_t = \{m, f, CoM_x, CoM_y, f_j\}$ with sufficient $\mu_t \in \mathbb{R}^{11L-1}$ and $\Sigma_t \in \mathbb{R}^{(11L-1) \times (11L-1)}$. 
The dual filter as shown in Fig.\ref{fig:framework} follows the structure of a Kalman filter with a \textit{prediction step} and an \textit{update step}, with the proposed novelty explained in this section.



\subsection*{Prediction Step}

\subsubsection{Constrained Monte Carlo Sampling}
\begin{figure}[h]
    \centering
    \includegraphics[width = \columnwidth]{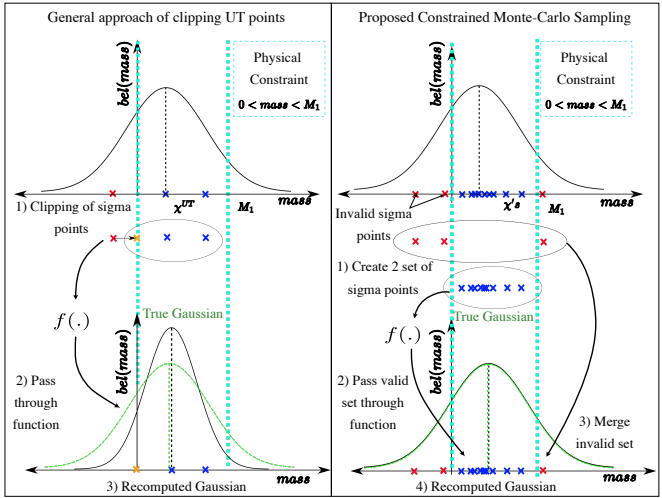}
    \caption{Illustration of the constrained Monte Carlo sampling compared with clipping of Sigma points}
    \label{fig:mc_sampling}
\end{figure}

In the prediction step, the next step joint belief is predicted given the prior belief and the actions. The object's inertial and interaction parameters are well-characterized physical quantities with some physical constraints (e.g. $m, f, f_j > 0$, $CoM_x$, $CoM_y$ must lie inside the object boundary). However, straightforward clipping the sigma points $\chi^{UT}$ in the UKF approach do not preserve the true variance of the Gaussian distribution \cite{ut_constraints}. Therefore, we present a novel constrained Monte Carlo sigma point sampling to preserve the physical constraints and the variance of the Gaussian \cite{dutta2023push}. A differentiable sampling method \cite{mc_method} was used to sample $C$ sigma points from the joint distribution $bel(\psi_{t-1}, \phi_{t-1})$ instead of Unscented Transformation: 
\begin{equation}
     \chi^{i}_{t-1} = \mu_{t-1} + \epsilon^{i} \sqrt{\Sigma_{t-1}}
     \label{eq:mc_sampling}
\end{equation}
where $i = \{1..C\}$ and $\chi_{t-1} = [\chi_{\psi_{t-1}}, \chi_{\phi_{t-1}}] \in \mathbb{R}^{C \times (11L-1)}$ and $\epsilon^{i} \sim \mathcal{N}(0, 1)$. Each sigma point has an associated weight $w_t^{i} = 1/C$. We set $C=100$ for all of our experiments. The sigma points undergo a filtering process to determine if they meet the physical constraints and are then fed into the data-driven model. However, even if the sigma points do not satisfy the constraints, they are still kept and reintroduced during the recomputation of the multivariate Gaussian in Eq. \ref{eq:predicted_belief} to maintain the uncertainty of the distribution. This process is depicted in Fig. \ref{fig:mc_sampling}. Subsequently, the filtered sigma points are utilized in the process model using Graph Neural Networks (GNN) and further elaborated as follows.

\subsubsection{Graph Generation \& Propagation}
\label{ssec:process}
To exploit the causal relationship and structure of the interaction, we utilize Graph Neural Networks \cite{sanchez2018graph} to model a novel representation of the object, the supporting surface, and the robot. This representation effectively captures the dynamics of the interaction following the principle of causality. Specifically, at every time step $t$, the graph is updated as the interaction forces transmitted from the robot to the links of the object (cause) and reciprocally to the robot (effect). Importantly, the resultant force on the robot encodes the necessary information for tactile observations. In addition, graph representation offers advantages in representing uniform, heterogeneous, and articulated objects with multiple links using the same structure and without making any additional changes to the neural network architectures.

Using the set of sigma points $\chi^{i}_{t-1} | i = \{1, .. C\}$ and the action $a_{t-1}$ of the robot, $C$ directed \textit{graph} $G^{i}_{t} = (\{\mathbf{n}_l\}_{l=1 .. L+2}, \{\mathbf{e}_j, s_j, r_j\}_{j=1 .. N_e})$ is created. $\{\mathbf{n}_l\}_{l=1 .. L+2}$ is a set of nodes where each $\mathbf{n}_l$ is a vector of node features. Node features comprise a structure similar to that of the state space, with dynamic (time-varying) factors i.e. the pose and twist values and static (time-invariant) factors i.e. the inertial parameters which are populated from the sigma points to the node features. The action parameters are used to populate the node that represents the robot, with default inertial parameters for the robot, and similarly for the table with zero pose and twist. 

A set of directed edges is created where $\mathbf{e}_j$ is created between the links of the object and the robot (the robot at any given time is interacting with a single link of the object), giving $2 \times (L-1) + 2$ edges and combing with the support edges representing the contact with support or table) for each link $N_e = 2 \times (L-1) + 2 + L = 3 \times L$ edges are generated. The static features of the edge are populated with frictional parameters from the sigma points. 
The cause and effect of the interaction are effectively represented by the directed edges, as they are involved in updating edge features over time. $s_j$ and $r_j$ are the indices of the sender and receiver nodes, respectively. We present an illustration of the graph generation and $1-step$ graph propagation of a 2 link \textit{articulated} object Fig. \ref{fig:graphnets}(a) undergoing a non-prehensile push interaction. 

To update the features of the nodes and the edges from time $t-1$ to $t$, we use a novel graph propagation algorithm described in Algorithm \ref{alg:gn}. The process of updating the graph network ($GP$) involves two main sub-functions, namely $f_n$ for nodes and $f_e$ for edges. In this case, we use three-layer feedforward networks to learn these functions. A single pass of the graph neural network can be seen as a step in message-passing on a graph \cite{sanchez2018graph}. We also illustrate propagation of the graph of an example articulated object for a single time step in Fig. \ref{fig:graphnets}(b). From the updated node features, we obtain the predicted sigma points:

\begin{align}
  \overline{\chi}_{\psi_t}&\longleftarrow GP(\chi_{t-1}, a_t) \\
   \overline{\chi}_{\phi_t} &= {\chi}_{\phi_{t-1}}
 \label{eq:process_model}
\end{align} 

$Q_t \in \mathbb{R}^{6 \times L}$ is the diagonal covariance process noise, which is user defined for a particular object and interaction type. The predicted next step sigma points $\overline{\chi}_t$, along with the process noise $Q_t$ are utilized to compute the expected Gaussian belief $\overline{bel}(\psi_t, \phi_t)$ as 
\begin{flalign}
    \overline{\chi}^{i}_{\psi_t} &= \overline{\chi}^{i}_{\psi_t} + \epsilon^{i} \sqrt{Q_t} \\
    \overline{\mu}_t &= \sum_{i=0}^{C}w_t^{i} \overline{\chi}_t \\
	\overline{\Sigma}_t &= \sum_{i=0}^{C}w_t^{i}(\overline{\chi}^{i}_t-\overline{\mu}_t)(\overline{\chi}^{i}_t-\overline{\mu}_t)^T
     \label{eq:predicted_belief}
\end{flalign} 
where, $i \in 1.. C$ and $\overline\chi_{t} = [\overline\chi_{\psi_{t}}, \overline\chi_{\phi_{t}}]$

\begin{algorithm}
  \caption{Graph Propagation Algorithm ($GP$)}
  \label{alg:gn}
  \textbf{Input}: Graph $G_{t-1}=(\{\mathbf{n}_i\}, \{\mathbf{e}_j, s_j, r_j\})$
  \begin{algorithmic}
  \State Initialize Stacks (LIFO)
  \State $NTV \longleftarrow \mathbf{n}_{R}$ \Comment{Nodes to visit}
  \State $NV  \longleftarrow \emptyset$ \Comment{Nodes visited}
  \State $EN \longleftarrow \emptyset$  \Comment{End nodes}
  \State \textcolor{blue}{\textit{Propagate cause}}
  \While $ \quad NTV \neq \emptyset$
    \State $\mathbf{n}_i = $ Pop $NTV$
    \State $\mathbf{n}_{{r}_{j}} = $ Gather receiver nodes of $\mathbf{n}_i$
    \State $\mathbf{n}_{{r}_{j}} = \mathbf{n}_{{r}_{j}} \setminus NV $ \Comment{Remove nodes already visited}
    \If{$\mathbf{n}_{{r}_{j}} \neq \emptyset$}
        \State Push $\mathbf{n}_i \rightarrow NV$
        \State Push $\mathbf{n}_{{r}_{j}} \rightarrow NTV$
        \For{each node $\mathbf{n}_{{r}_{j}}$}
            \State Compute causal edges, $\mathbf{e}^{*}_{j} = f_e(\mathbf{n}_{i}, \mathbf{n}_{{r}_{j}}, \mathbf{e}_{s_{j}})$  
            \State \Comment{$\mathbf{e}_{s_{j}}$ is static edge feature (friction values)}
             \State Compute support edges, $\mathbf{e}^{*}_{k} = f_e(\mathbf{n}_{S}, \mathbf{n}_{{r}_{j}}, \mathbf{e}_{s_k})$
            \State Compute node features, $\mathbf{n}^{*}_{i} = f_n(\mathbf{n}_{i}, \mathbf{e}^{*}_{j}+\mathbf{e}^{*}_{k})$
        \EndFor
    \Else
        \State Push $\mathbf{n}_i \rightarrow EN$
    \EndIf
    
  \EndWhile
  \State \textit{Propagate effect}
  \While $\quad NV \neq \emptyset$
    \State $\mathbf{n}_i = $ Pop $NV$
    \State $\mathbf{n}^{*}_{{s}_{j}} = $ Gather sender nodes of $\mathbf{n}_i$
    \State $\mathbf{n}^{*}_{{s}_{j}} = \mathbf{n}^{*}_{{s}_{j}} \setminus NV$
     \State Aggregate effect edges, $\mathbf{e}^{*}_{j} = f_e(\mathbf{n}^{*}_{{s}_{j}}, \mathbf{n}_{i},  \mathbf{e}_{s_{j}})$
     \State Update node features, $\mathbf{n}^{*}_{i} = f_n(\mathbf{n}_{i}, \sum_{j/s_j}\mathbf{e}^{*}_{j})$
 \EndWhile
  \end{algorithmic}
  \textbf{Output}: Graph $G_{t}=(\{\mathbf{n}^{*}_i\}, \{\mathbf{e}^{*}_j, s_j, r_j\})$
\end{algorithm}


\subsection*{Update Step}
The dual filter employs a separate update of parameter belief similar to the parameter update presented in \cite{liu_west} and the conditional pose belief update based on the UKF update \cite{prob_rob_book}.  

To update the joint belief, we require an observation model to predict the observation sigma points $\overline{\mathcal{Z}}_t$, which must account for visual and tactile observations. To reduce the complexity of predicting raw RGB-D images, we use the initial segmented point cloud $\mathcal{PC}_{t_0}$ from the shape perception method to transform it using the predicted pose and generate expected RGB-D images using the standard 3D to 2D projective transformation approach \cite{nematollahi2022t3vip} involving the intrinsic and extrinsic values of the camera. This overcomes the generalization problem faced by synthetic visual networks as in \cite{dutta2023push} and can be used for any novel object. For the tactile counterpart, a three-layer feedforward network is utilized to predict the contact force information from the edge encoding directed towards the robot. The predicted observation sigma points $\overline{\mathcal{Z}}_t$ are given by:

\begin{flalign}
     \overline{\mathcal{Z}}^{V}_t &= \mathbf{w}(\overline{\chi}^{'}_{{\psi}_t}, \mathcal{PC}_{t_0}) \\
     \overline{\mathcal{Z}}^{T}_t &\longleftarrow  TacNet(\overline{\chi}^{'}_t, enc_{a_t})\\
     R_t &\longleftarrow ObsNoiseNet(z^V_t, z^T_t)
    \label{eq:observation_model}
\end{flalign}

\noindent where $\mathbf{w}$ is the projective transformation function. The observation noise model $ObsNoiseNet$ uses three-layer CNN to predict heteroscedastic visual noise $\sigma^{{obs}^{V}}_t \in \mathbb{R}$ and a two-layer feedforward network to predict heteroscedastic tactile noise $\sigma^{{obs}^{T}}_t \in \mathbb{R}$. This is used to construct the diagonal observation noise matrix $R_t \in \mathbb{R}^{4098 \times 4098}$:
\begin{equation}
R_t = diag [\underbrace{\sigma^{{obs}^{V}}_t, \cdots \sigma^{{obs}^{V}}_t}_{4096} , \sigma^{{obs}^{T}}_t, \sigma^{{obs}^{T}}_t]
\end{equation}

The RGB-D images $o^{V}_t$ are transformed into grayscale and resized to $64 \times 64$ size. They are then flattened and merged with tactile observations $o^{T}_t$ to create $z_t \in \mathbb{R}^{4098}$, which is used for the update process.


\subsubsection*{Parameter Update}
We update the weights based on the likelihood of the observation sigma points $\overline{\mathcal{Z}}_t = [\overline{\mathcal{Z}}^{T}_t, \overline{\mathcal{Z}}^{V}_t] $ in the observation distribution $\sim \mathcal{N}(.|z_t, R_t)$
\begin{flalign}
    w_t^{j} &=  w_t^{j}e^{\big(-\frac{1}{2}(\overline{\mathcal{Z}}^{j}_t - z_t) R^{-1}(\overline{\mathcal{Z}}^{j}_t - z_t)^T)}
     \label{eq:parameter_update}
\end{flalign}
where $j \in 1, .. C$. The updated parameter belief $bel(\phi_t)$ is recomputed  via a Gaussian Smooth Kernel \cite{liu_west} method after normalizing the updated weights:
\begin{flalign}
    \mu_{\phi_t} &= \sum_{i=0}^{C}w_t^{[i]} \overline{\chi}^{'}_{\phi_t};  \quad m^{[i]}_{\phi_t} = a\overline{\chi}^{'}_{{\phi}_t}+(1-a)\mu_{\phi_t} \\
	\Sigma_{\phi_t} &= h^2\sum_{i=0}^{C}w_t^{[i]}m^{[i]}_{\phi_t}-\mu_{\phi_t}
\end{flalign}
where $a$ and $h=\sqrt{1-a^2}$ are shrinkage values of the kernels that are user-defined and set to 0.01, and $m$ are the kernel locations.

\subsubsection*{Pose Update}
We make use of the dependence of the pose on the parameters to compute the conditional pose distribution $bel(\psi_t | \phi_t) \sim \mathcal{N}(\psi_t|\mu_{\psi_t|\phi_t}, \Sigma_{\psi_t|\phi_t})$ using the Multivariate Gaussian Theorem \cite{mult_gauss}
\begin{flalign}
    \mu_{\psi_t|\phi_t} &= \psi_t + \Sigma_{\psi_t\phi_t}\Sigma^{-1}_{\phi}(\phi_t - \mu_{\phi_t})\\
    \Sigma_{\psi_t|\phi_t} &= \Sigma_{\psi_t} -  \Sigma_{\psi_t\phi_t}\Sigma^{-1}_{\phi_t}\Sigma_{\phi_t\psi_t}
    \label{eq:cond_pose_distribution}
\end{flalign}

For the update of the conditional pose, the standard Unscented Kalman Filter (UKF) \cite{prob_rob_book} is used on the predicted conditional pose distribution $\overline{bel}(\psi_t | \phi_t = \mu_{\phi_t})$ using Eq.\ref{eq:cond_pose_distribution}. The $\mu_{\phi_t}$ of the updated parameter belief is used with the predicted pose sigma points $\overline{\chi}^{UT}_{\psi_t}$ to obtain the predicted observation sigma points $\overline{\mathcal{Z}}'_t$. After the conditional pose update, the posterior joint is computed as
\begin{equation}
     bel(\psi_t, \phi_t) = bel(\psi_t|\phi_t)bel(\phi_t)
     \label{eq:belief_joint_update}
 \end{equation}
Note that the cross-covariance matrices $\Sigma_{\psi_t\phi_t}, \Sigma_{\phi_t\psi_t}$ are not updated through the dual update step and are kept constant. The updated posterior joint belief is used as a prior to filter, then as the next time step. The filtering step is used end-to-end for both learning and inference. In the following section, we present the experimental setup and the results obtained. 

%% file: sections/experiments.tex
\section{Experiments}
\label{sec:exp}
\begin{figure*}[!htb]
    \centering
    \includegraphics[width = \textwidth]{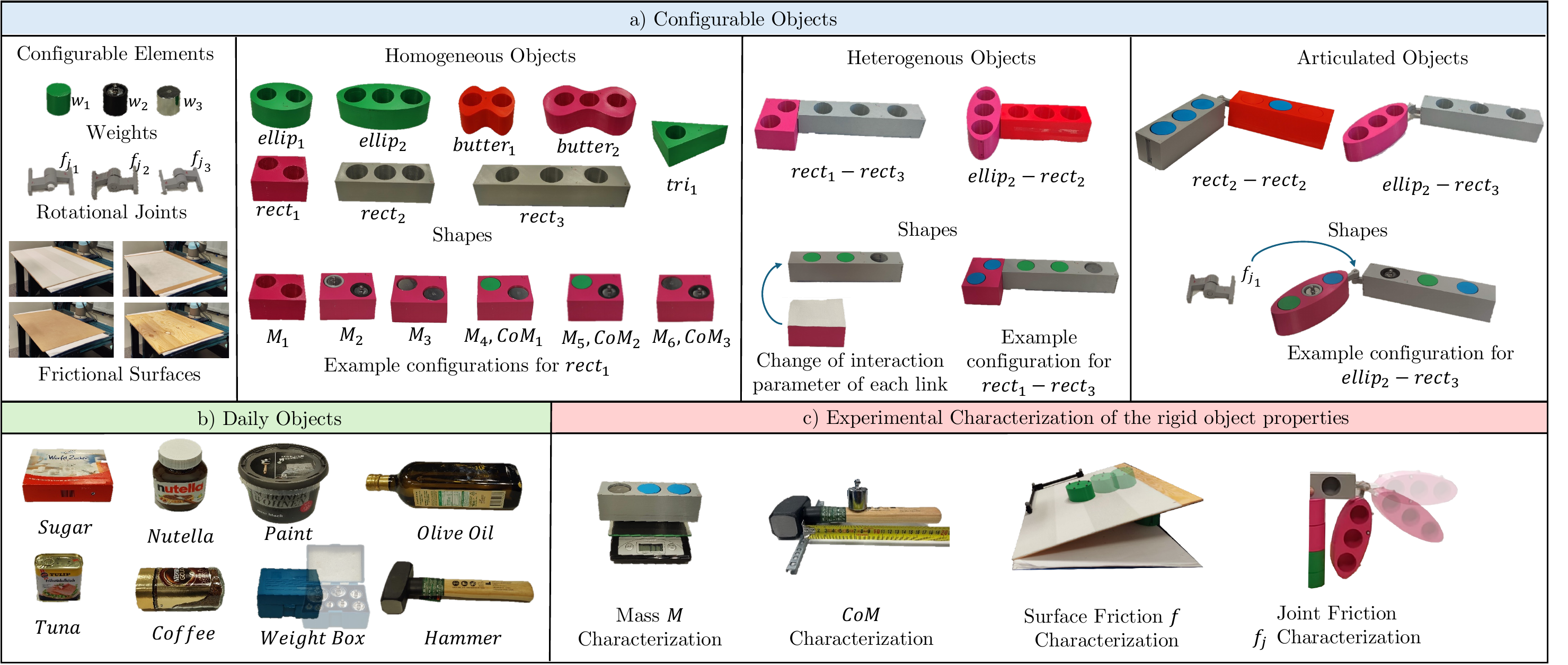}
    \caption{Experimental Setup utilized to validate the framework. a) Presents the configurable 3D printed object designed to be used as diverse objects of types: \textit{homogeneous}, \textit{heterogeneous}, \textit{articulated}. b) Presents the daily objects used to validate the generalization of our approach c) Illustrates the experimental characterization methods employed to measure the ground truth ($GT$) inertial and frictional properties.} \label{fig:experimental_object_set}
\end{figure*}

Here, we provide an explanation of the experimental setup and the validation experiments performed. To our knowledge, no previous approach addressed such a diverse range of object sets - \textit{homogeneous}, \textit{heterogeneous}, and \textit{articulated} using a single framework. Therefore, we present a comprehensive comparison between our proposed work, referred to as $A-GNN$  and the previous approach in \cite{dutta2023push}, which is referred to as the baseline $A-FF$. The previous approach used feedforward networks instead of graph networks that require additional assumptions about the number of links present in the case of \textit{heterogeneous} and \textit{heterogeneous} objects. For our comparative experimental study, we have updated the visual observation model of $A-FF$, which used a synthetic sensor model with the current approach of using a 3D projective transformation to improve generalization to new objects and without requiring additional training of the synthetic visual network. 

Furthermore, we have implemented and compared an analytical model of non-prehensile pushing \cite{lynch1992manipulation} for pose and parameter inference of \textit{homogeneous} objects, which we will refer to as baseline $A-Analytical$. The analytical model is described in detail in the Appendix \ref{app:anal_model} for reference. Both $A-FF$ and $A-Analytical$ utilized the dual filtering setup with active action affordance selection, with differences in the process and observation models compared to $A-GNN$. In addition, the process and observation noise for the analytical model were manually adjusted and fixed, to additionally evaluate the effect of the learned noise models.

We also performed ablation studies to compare the active action selection strategy with uniform action selection ($U-GNN$) and random action selection ($R-GNN$) for both model learning and inference. In addition, we present the results of the application of our proposed framework for pose estimation, goal-driven control, and environmental change detection.

\subsection{Experimental Setup}
\label{subsec:experiment_setup}
In this work, we used a real robotic setup to validate the proposed approach and compare it with the baselines. The robotic setup consists of a Universal Robots UR5 with a Robotiq two-finger gripper and a Franka Emika Panda robotic manipulator as shown in Figure \ref{fig:problem_setup}. Two Contactile tactile sensors \cite{contactile} were attached to the inner and outer surfaces of the robotiq gripper finger pad, and a Zed2i stereo camera \cite{zed} was rigidly attached to the Panda arm. The maximum speed allowed for UR5 and Panda was 25 $mm/s$ due to safety constraints. The ground truth values of the pose were collected using the motion capture system - Optitrack \cite{OptiTrac66:online}. 

We designed and developed 3D printed objects that can be configured by adjusting their physical parameters, using weights, different supporting frictional surfaces, and different joints of varying friction for articulated objects. This allowed us to generate a wide range of objects, including those that are \textit{homogeneous}, \textit{heterogeneous}, and \textit{articulated}. These objects are depicted in Fig. \ref{fig:experimental_object_set}(a). A total of 120 object configurations were selected that ensured a sufficient variation of physical properties (40 \textit{homogeneous}, 40 \textit{heterogeneous}, 40 \textit{articulated}). In addition to configurable 3D printed objects, a few everyday objects with distinct physical properties were selected for the validation of the real-robotic use case, as shown in Fig. \ref{fig:experimental_object_set}(b).

The objects were experimentally characterized as depicted in Fig. \ref{fig:experimental_object_set}(c) and the measured physically parameters are referred to as ground-truth ($GT$). To determine the friction value between the surface and each object, the supporting surface was tilted and the critical angle at which the object started to slide, was measured to determine the coefficient of kinetic friction \cite{bala_21}. Furthermore, for articulated objects, joint friction was characterized by releasing the connected link as a pendulum and following the angle of the joint \cite{martin2022coupled} using the motion capture setup. For \textit{heterogeneous} objects, $GT$ $\mu(f_j)$ was set to $1$ (rigid joint), to handle both \textit{heterogeneous} and \textit{articulated} objects autonomously, having a consistent state space.


\subsection{Experimental Results}
\label{subsec:experiment_results}
This section presents details of the various experiments performed in the robotic setup to validate the proposed framework.

\subsubsection{Active Shape Perception}
\begin{figure}[t!]
    \centering
    \includegraphics[width = \columnwidth]{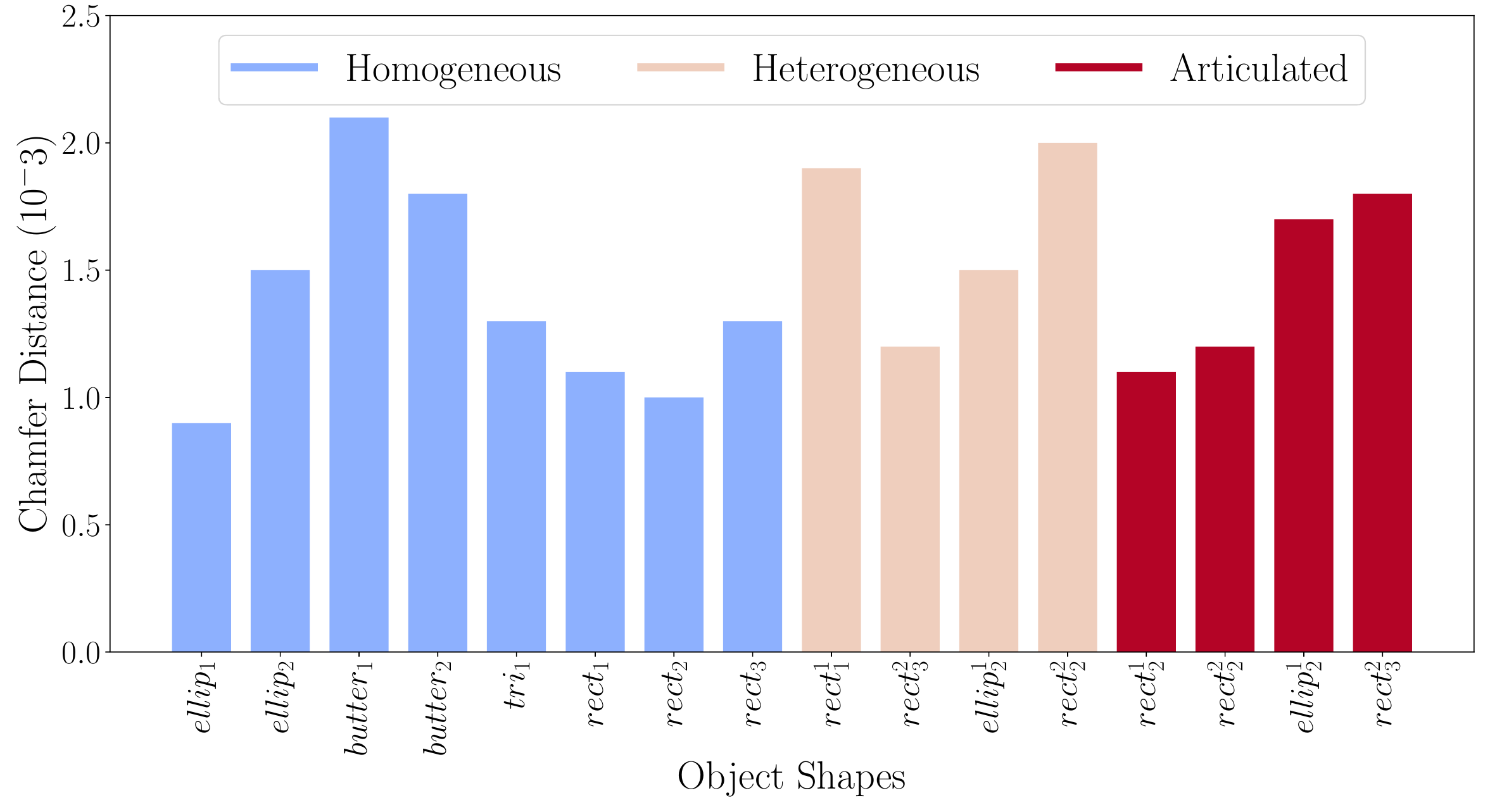}
    \caption{Chamfer Distance of the estimated superquadrics with respect to ground truth object shape. Please see Fig.\ref{fig:experimental_object_set} for visualization of the object shapes corresponding to the annotated shape name. The labels $rect^{1}_1$ and $rect^{2}_3$ in $heterogeneous$ section denote a combined object with $rect^{1}_1$ as the first link and $rect^{2}_3$ denotes the second link.}
\label{fig:shapeplot}
\end{figure}

We present quantitative results on the estimation of the shapes of the various configurable objects, as the ground truth ($GT$) shape of these objects was present. Figure \ref{fig:shapeplot} presents the bar graph of the Chamfer distance ($CD$) metric of the estimated shape after multiple active views with respect to the $GT$ shape. The lower $CD$ represents a better estimation of the shape by the superquadric. In general, superquadrics are capable of accurately representing the different shapes of objects used in the study (with a maximum value $CD$ of 0.002 for $butter_1$). It should be noted that when the same shape, such as $rect_3$, is used as part of a \textit{heterogeneous} object, it exhibits a higher $CD$ compared to \textit{homogeneous} and \textit{articulated} objects. This is expected since the various links in the case of \textit{heterogeneous} objects are positioned close to each other.

In addition, Fig.\ref{fig:activeshape} illustrates the improvement of the estimated superquadric with an increasing number of views (lower $CD$). The plot compares the proposed active next-best view approach which can consistently converge in $4$ views compared to uniform $6$ and random $5$ next-best view selection for shape perception.

\begin{figure}[!htb]
    \centering
    \includegraphics[width = 0.8\columnwidth]{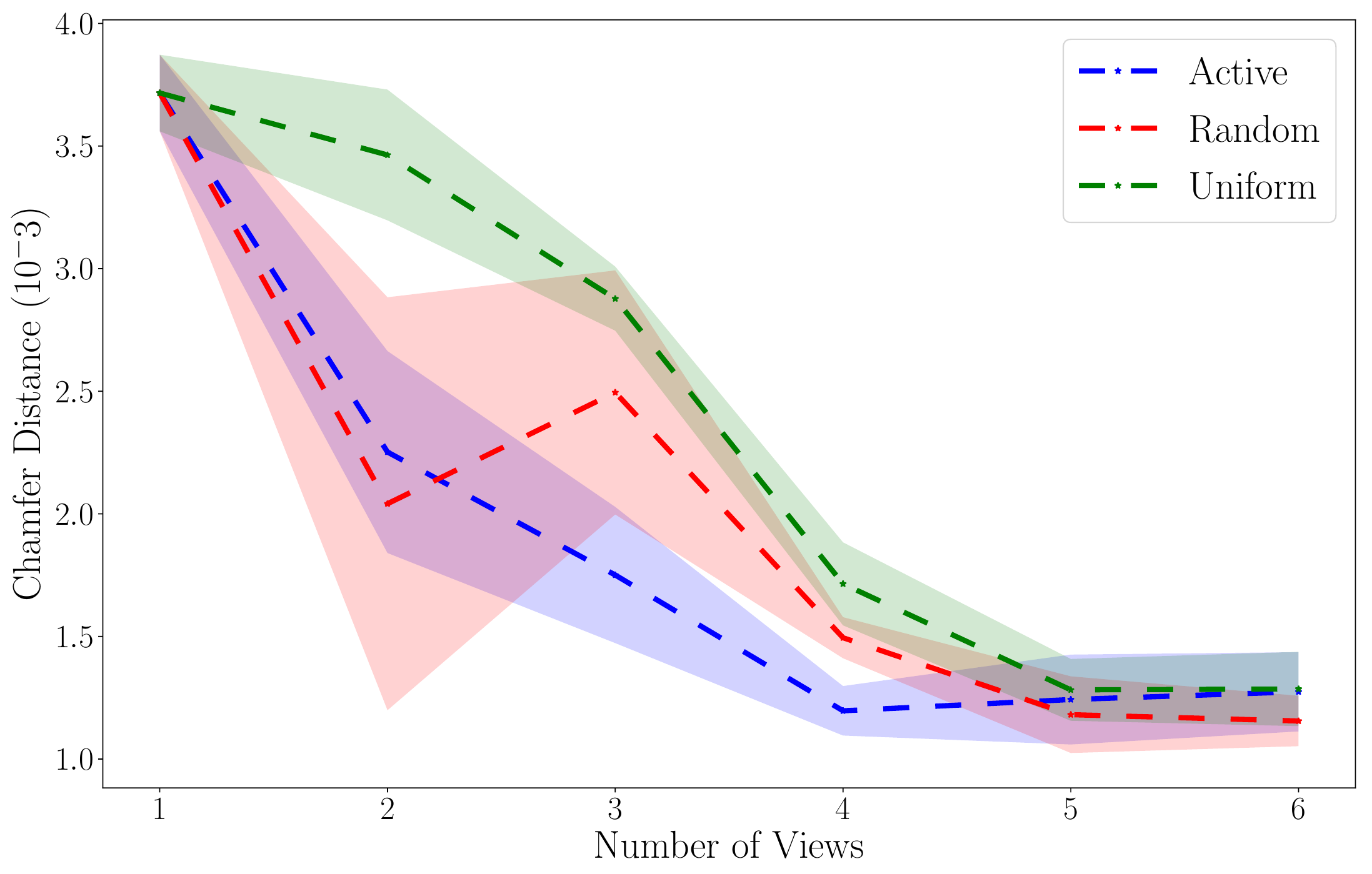}
    \caption{Comparison of Active, Uniform and Random shape perception result with standard deviation with iterative} views with Chamfer Distance as metric.
    \label{fig:activeshapeplot}
\end{figure}

We also present a qualitative result of shape perception, showing the visual accuracy of the estimated superquadrics, for all the different shapes of objects used in the work in Fig. \ref{fig:qualshapeplot}. presented in the Appendix \ref{app:qualshape}.

\subsubsection{Learning of Dual Differentiable Filter}
To train the networks in the differentiable filter, we leveraged a combination of negative log-likelihood loss ($\mathcal{L}_{NLL}$) \cite{alina_2_main}, mean squared error loss ($\mathcal{L}_{MSE}$), and observed noise log-likelihood loss ($\mathcal{L}^{obs}_{NLL}$). The $\mathcal{L}_{NLL}$ and $\mathcal{L}_{MSE}$ were calculated using the pose, parameter and contact forces of the ground truth ($GT$) pose, parameter, and contact forces with respect to the joint belief and the predicted contact force distribution. The $\mathcal{L}^{obs}_{NLL}$ was used to improve the visual observation noise model training and was calculated as the log-likelihood of projected images using ground truth pose information and observed visual images with associated observation noise. For each type of interaction, a separate model was trained because of their distinct dynamics.


The time horizon for each interactive action was set at $t_H=5$ seconds, with a sampling rate of $15$ Hz.  In the active approach, the $N$-step look-ahead parameter of $N=2.5$ seconds was used. Of the 120 configurable objects, 80\% was used for training, 10\% for validation, and 10\% for testing purposes. We employed iterative training \cite{ren2021survey} with the Adam optimizer until the validation loss reached convergence. In the active case, iterative training involved executing the top five informative actions sampled, which were then added to the existing trajectory buffer. This buffer ensured the replacement of initial trajectories, followed by multiple epochs over the buffer until the validation loss converged. Each action trajectory was further split into three to allow for a substantial batch size. This step was repeated when the pose and contact force prediction error in the validation set reached a minimum threshold. 

\begin{figure}[!htb]
    \centering
    \includegraphics[width = \columnwidth]{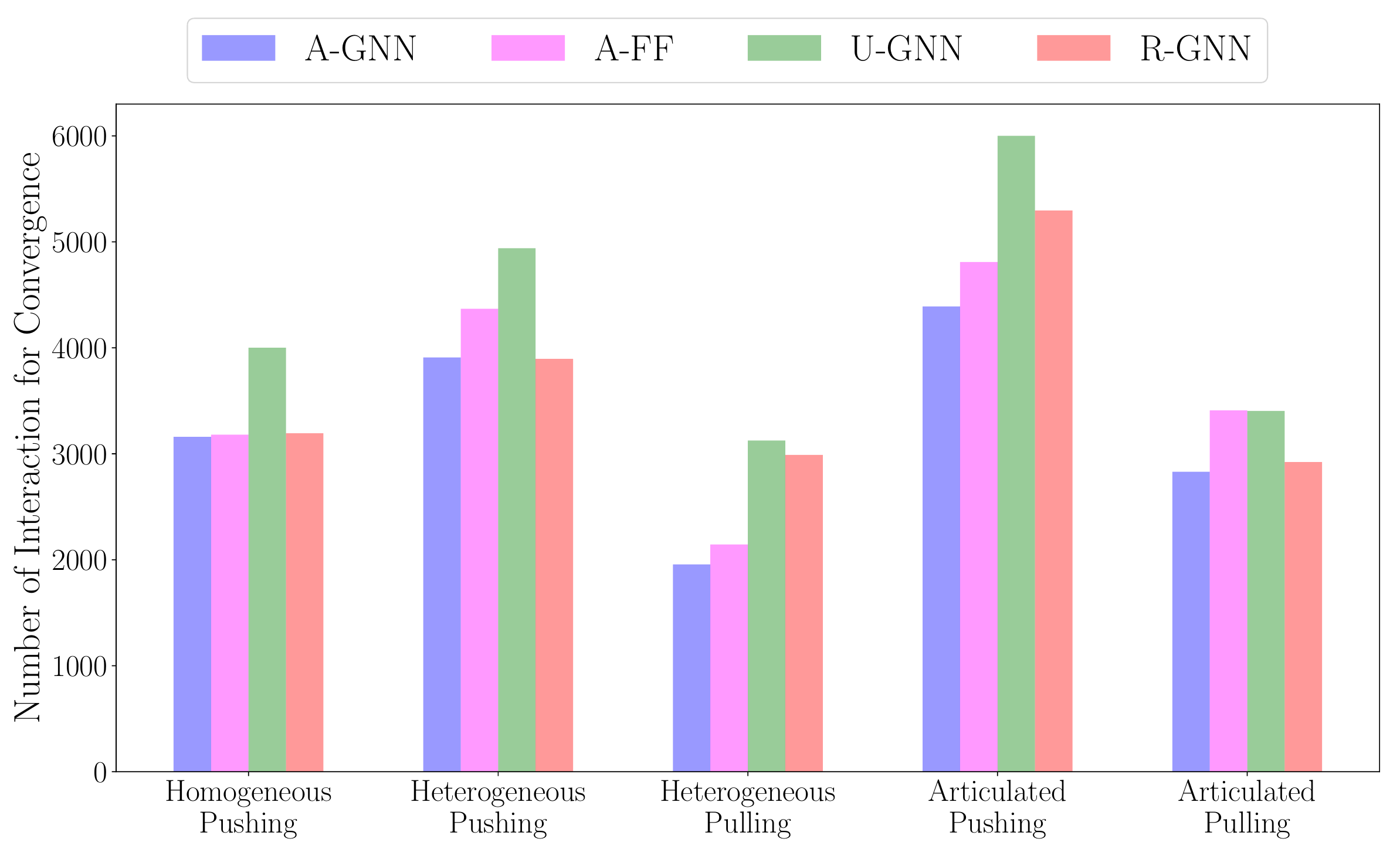}
    \caption{Results on number of interaction required for training the models for the proposed $A-GNN$ and baseline $A-FF$, $U-GNN$, $R-GNN$ for different object types \textit{Homogeneous}, \textit{Heterogeneous} and \textit{Articulated} objects with \textit{pushing} and \textit{pulling}.}
    \label{fig:learningplot}
\end{figure}

Fig. \ref{fig:learningplot} illustrates the comparison of the number of required trajectories needed to train models for the \textit{homogeneous}, \textit{heterogeneous}, and \textit{articulated} object scenarios, considering both non-prehensile pushing and prehensile pulling interactions. The results obtained using the $A-GNN$ and $A-FF$ approaches involved active actions for learning. On the other hand, $U-GNN$ and $R-GNN$ represent a uniform and random action selection strategy for learning in the proposed model instead of active actions. It can be observed that the $A-GNN$ approach consistently requires the least number of interactions for training. Additionally, even when active actions are used, the $Active-FF$ approach requires more interactions for the \textit{heterogeneous} and \textit{articulated} cases, thus highlighting the superiority of GNNs over feedforward networks for learning the dynamics of complex interactions. 

\begin{figure}[tb!]
    \centering
    \includegraphics[width = \columnwidth]{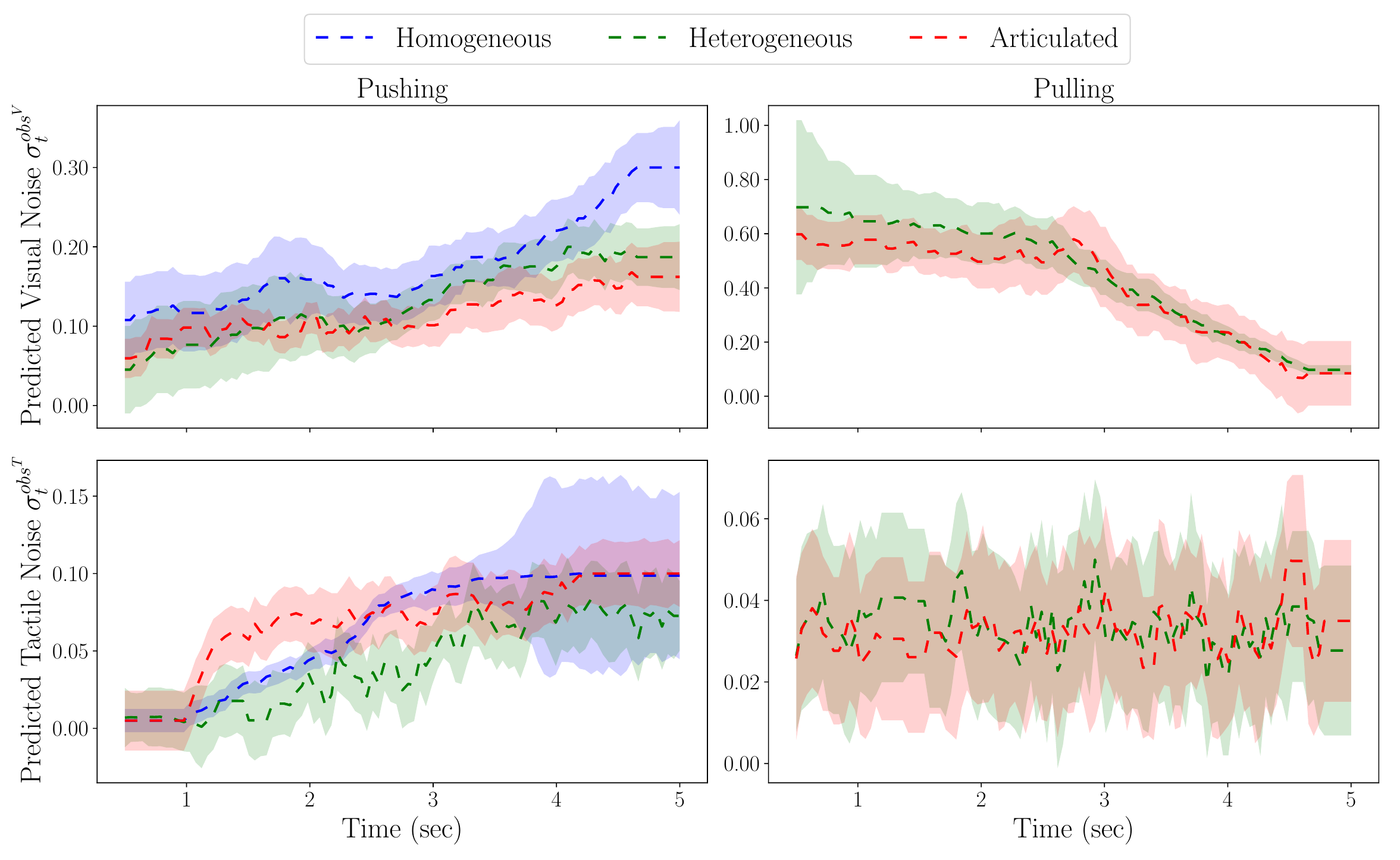}
    \caption{Presents the prediction noise variance from the learned noise models for the validation set of different object and interaction types.}
    \label{fig:learned_noise_models}
\end{figure}

In addition, we present the observation noise prediction for various objects during the non-prehensile pushing and prehensile pulling interactions at each time step in the validation set in Fig.\ref{fig:learned_noise_models}. We can observe the heteroscedastic nature of both visual and tactile noise in our robotic setup, i.e. the noise present in the observations is not constant but depends on the state (e.g. occlusion due to robot or loss in contact with the object). The results in Fig.\ref{fig:learned_noise_models} suggest that tactile noise is much lower than visual noise, especially in the case of prehensile interaction, making tactile sensing modality more reliable for inference. 

\subsubsection{Parameter Inference}
For parameter inference of unknown (test) objects, we performed multiple interactive actions. At the end of each, the posterior belief of the object parameters was used to initialize the belief for the next action. As the different inertial and interaction parameters have different ranges and units (mass in $kg$, $CoM$ in $m$), we report the estimation error using the normalized root mean squared error $NRMSE$ \cite{dutta2023push} between the predicted mean value of the parameter and the ground truth value. The normalization factor for each parameter was calculated as the maximum range of values in the object set. Furthermore, due to the symmetric nature of the objects, we compute $CoM$ as the radial distance from the geometric from the estimated $CoM_x$ and $CoM_y$ 

\begin{figure}[!tbh]
    \centering
    \includegraphics[width = \columnwidth]{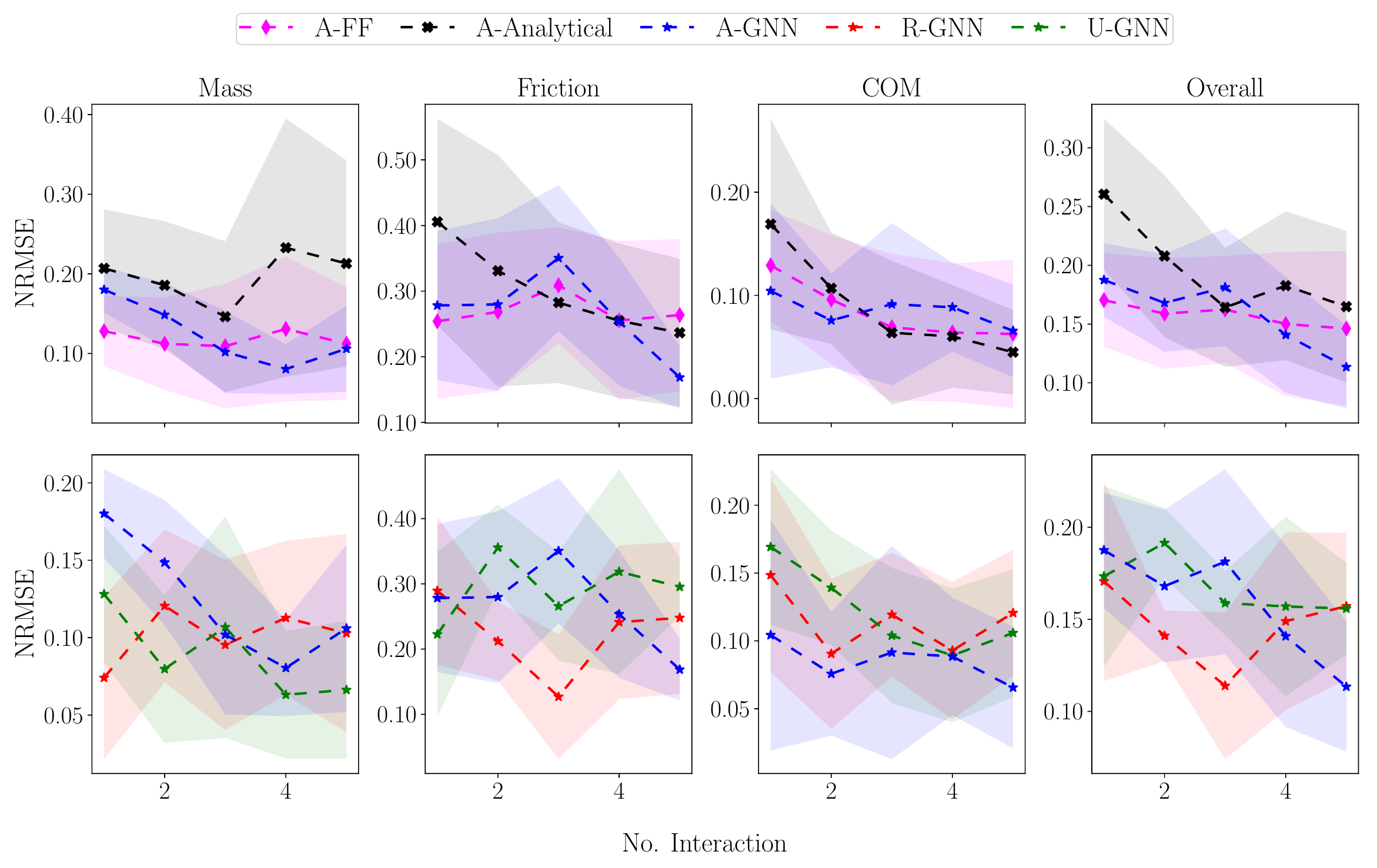}
    \caption{Parameter estimation error across multiple interactions for homogeneous objects with \textit{pushing} interaction comparing proposed $A-GNN$ with $A-FF$ and $A-Analytical$ in top row and ablative action selection $U-GNN$} and $R-GNN$ in the bottom row
    \label{fig:inferencehomo}
\end{figure}

\begin{figure}[!tbh]
    \centering
    \includegraphics[width = \columnwidth]{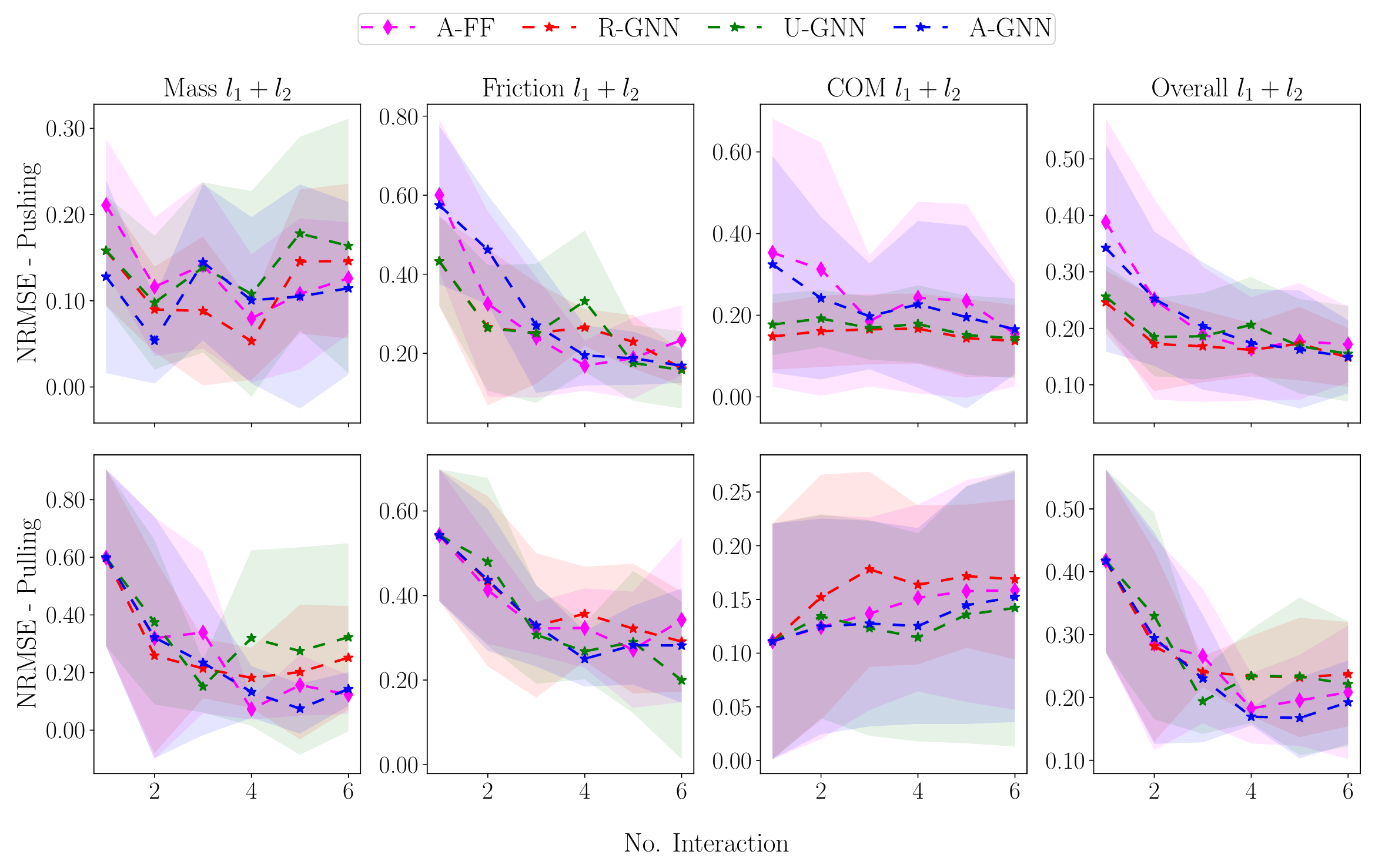}
    \caption{Parameter estimation error across multiple interactions for heterogeneous objects comparing proposed $A-GNN$ with $A-FF$, $R-GNN$ and $U-GNN$. The top row presents the errors for \textit{pushing} interaction and the bottom row for \textit{pulling} interaction}
    \label{fig:inference_hetero}
\end{figure}

\begin{figure}[!tbh]
    \centering
    \includegraphics[width = \columnwidth]{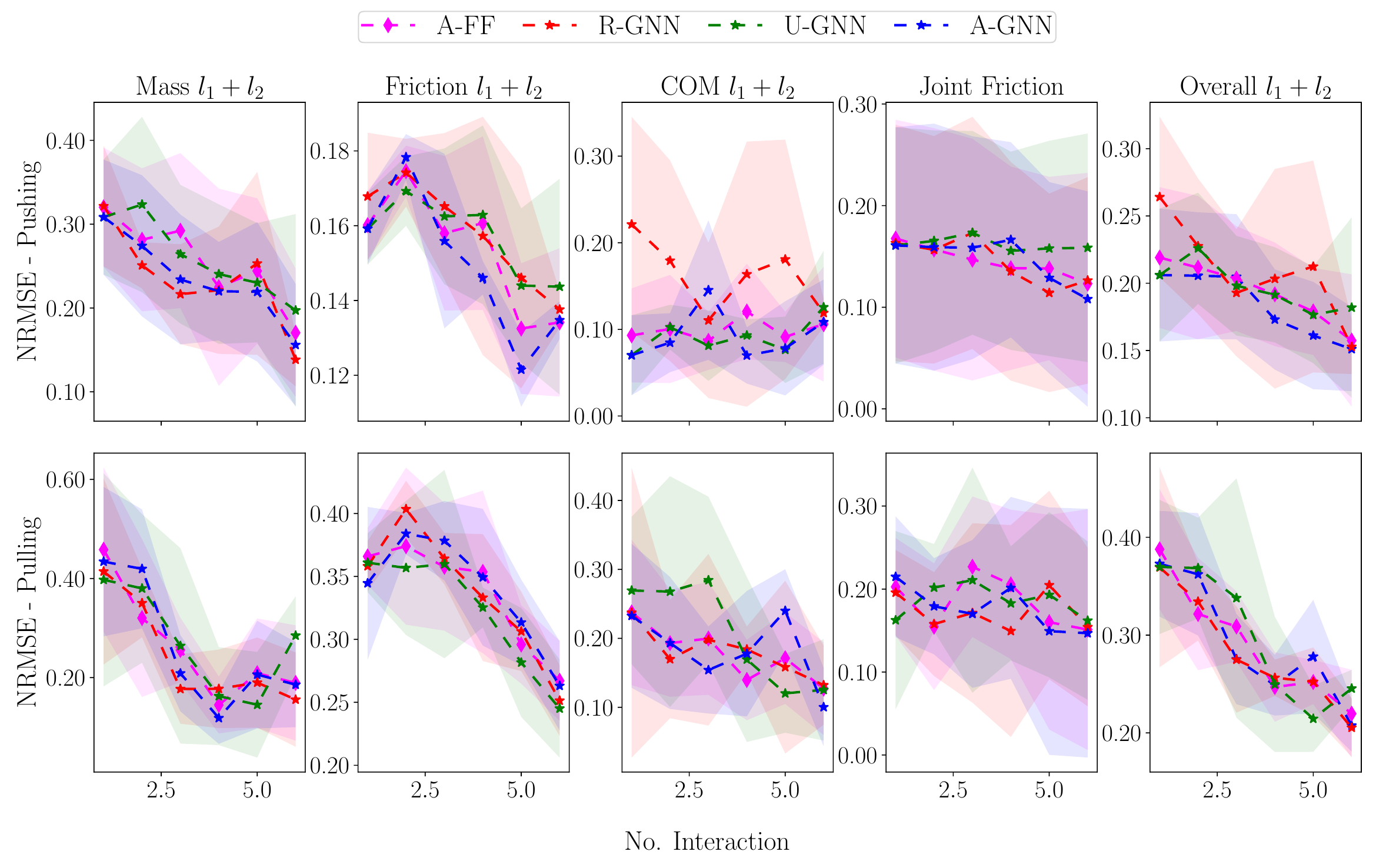}
    \caption{Parameter estimation error across multiple interactions for articulated objects comparing proposed $A-GNN$ with $A-FF$, $R-GNN$ and $U-GNN$. The top row presents the errors for \textit{pushing} interaction and the bottom row for \textit{pulling} interaction}
    \label{fig:inference_articulated}
\end{figure}

To understand the effect of multiple links and explore the robustness of the inference step, we chose to interact only with the largest link in the case of \textit{articulated} and \textit{heterogeneous} objects which is denoted as $l_1$, and the subsequent link as $l_2$. The results of the parameter estimation for all test objects after every interaction are presented in Fig. \ref{fig:inferencehomo} for \textit{homogeneous} objects, Fig.\ref{fig:inference_hetero} for \textit{heterogeneous} and Fig.\ref{fig:inference_articulated} for \textit{articulated} objects. We compare the parameter estimation error of the $A-GNN$ approach with the baselines $A-Analytical$ (only for \textit{homogeneous}) objects as well as $A-FF$. In addition to the baselines, we also present the ablation study of the impact of active action on parameter inference and compare it with $U-GNN$ and $R-GNN$. 

We can observe that as the complexity of objects increases from \textit{homogeneous}, to \textit{heterogeneous} and \textit{articulated}, more interactions are required for convergence. The various properties are coupled during the interaction, resulting in a higher standard deviation, which can be especially observed in \textit{ heterogeneous} objects and the estimation of joint friction $f$. In addition, it can be seen from Figs. \ref{fig:inference_hetero} and \ref{fig:inference_articulated}, that the \textit{pulling} interaction has a lower $NRMSE$ along with a lower standard deviation, compared to the \textit{pulling} interaction.

Furthermore, Tables \ref{tab:inference_homo}, \ref{tab:inference_hetero}, \ref{tab:inference_articulated} present the final error values after the variance in parameters has been reduced after multiple interactions. The tabulated results demonstrate that the proposed $A-GNN$ consistently estimates the properties of objects with the least $NMRSE$. Although $R-GNN$ performs similarly in the case of $pushing$ interaction, the performance deteriorates during \textit{pulling}. 


\begin{table}[!tbh]
\caption{Tabulated result of the individual parameter estimation error for \textit{homogeneous} objects when the variance over the parameters converged after multiple interactions}
\label{tab:inference_homo}
\resizebox{\columnwidth}{!}{%
\begin{tabular}{l|l|l|l|l|l|}
                              & \multicolumn{1}{c|}{A-GNN} & \multicolumn{1}{c|}{U-GNN} & \multicolumn{1}{c|}{R-GNN} & \multicolumn{1}{c|}{A-Analytical} & \multicolumn{1}{c|}{A-FF} \\ \hline
\multicolumn{1}{|l|}{$m$}     & 0.11 $\pm$0.05             & \textbf{0.07 $\pm$ 0.04}   & 0.10 $\pm$ 0.06            & 0.21 $\pm$ 0.13                   & 0.11 $\pm$ 0.07           \\ \cline{1-1}
\multicolumn{1}{|l|}{$f$}     & \textbf{0.17 $\pm$ 0.05}   & 0.29 $\pm$ 0.05            & 0.24 $\pm$ 0.12            & 0.23 $\pm$ 0.11                   & 0.26 $\pm$ 0.12           \\ \cline{1-1}
\multicolumn{1}{|l|}{$CoM$}   & 0.07 $\pm$0.05             & 0.11 $\pm$ 0.05            & 0.12 $\pm$ 0.05            & \textbf{0.04 $\pm$ 0.03}          & 0.06 $\pm$ 0.07           \\ \cline{1-1}
\multicolumn{1}{|l|}{Overall} & \textbf{0.11 $\pm$ 0.04}   & 0.15 $\pm$ 0.02            & 0.16 $\pm$ 0.04            & 0.16 $\pm$ 0.06                   & 0.15 $\pm$ 0.06           \\ \hline
\end{tabular}%
}
\end{table}

\begin{table*}[]
\centering
\caption{Tabulated result of the individual parameter estimation error for \textit{heterogeneous} objects with 2 links when the variance over the parameters converged after multiple interactions. The exploratory action is applied to $l_1$. For each interaction \textit{pulling} and \textit{pushing}, the best performing approach is highlighted in bold face.}

\label{tab:inference_hetero}
\resizebox{\textwidth}{!}{%
\begin{tabular}{l|ll|ll|ll|ll|}
                                    & \multicolumn{2}{c|}{A-GNN}                                                   & \multicolumn{2}{c|}{U-GNN}                                                   & \multicolumn{2}{c|}{R-GNN}                                                   & \multicolumn{2}{c|}{A-FF}                                                    \\ \cline{2-9} 
                                    & \multicolumn{1}{c}{\textit{Pushing}} & \multicolumn{1}{c|}{\textit{Pulling}} & \multicolumn{1}{c}{\textit{Pushing}} & \multicolumn{1}{c|}{\textit{Pulling}} & \multicolumn{1}{c}{\textit{Pushing}} & \multicolumn{1}{c|}{\textit{Pulling}} & \multicolumn{1}{c}{\textit{Pushing}} & \multicolumn{1}{c|}{\textit{Pulling}} \\ \cline{1-1}
\multicolumn{1}{|l|}{$m_{l_1}$}     & \textbf{0.11 $\pm$ 0.10}             & 0.14 $\pm$ 0.06                       & 0.16$\pm$0.15                        & 0.32 $\pm$ 0.12                       & 0.15$\pm$0.09                        & 0.25 $\pm$ 0.18                       & 0.13 $\pm$ 0.06                      & \textbf{0.12 $\pm$ 0.07}              \\ \cline{1-1}
\multicolumn{1}{|l|}{$m_{l_2}$}     & \textbf{0.11 $\pm$ 0.10}             & 0.14 $\pm$0.05                        & 0.16$\pm$0.15                        & 0.32 $\pm$ 0.13                       & 0.15$\pm$0.09                        & 0.25 $\pm$ 0.18                       & 0.13 $\pm$ 0.06                      & \textbf{0.12 $\pm$ 0.07}              \\ \cline{1-1}
\multicolumn{1}{|l|}{$f_{l_1}$}     & \textbf{0.12 $\pm$ 0.12}             & 0.4 $\pm$0.16                         & 0.17$\pm$0.14                        & \textbf{0.26 $\pm$0.17}               & 0.22$\pm$0.11                        & 0.33 $\pm$ 0.19                       & 0.20 $\pm$ 0.11                      & 0.32 $\pm$ 0.19                       \\ \cline{1-1}
\multicolumn{1}{|l|}{$f_{l_2}$}     & 0.21 $\pm$ 0.11                      & 0.16 $\pm$ 0.14                       & 0.15$\pm$0.09                        & \textbf{0.14 $\pm$ 0.14}              & \textbf{0.10$\pm$0.06}               & 0.25 $\pm$ 0.15                       & 0.26 $\pm$ 0.16                      & 0.36 $\pm$ 0.21                       \\ \cline{1-1}
\multicolumn{1}{|l|}{$CoM_{l_1}$}   & 0.17 $\pm$0.16                       & 0.17 $\pm$ 0.12                       & 0.19$\pm$0.12                        & \textbf{0.17 $\pm$ 0.11}              & 0.18$\pm$0.12                        & 0.18 $\pm$ 0.12                       & \textbf{0.16 $\pm$ 0.15}             & 0.18 $\pm$ 0.20                       \\ \cline{1-1}
\multicolumn{1}{|l|}{$CoM_{l_2}$}   & 0.16 $\pm$0.12                       & 0.13 $\pm$ 0.10                       & \textbf{0.09$\pm$0.06}               & \textbf{0.11 $\pm$ 0.07}              & 0.10$\pm$0.09                        & 0.16 $\pm$ 0.08                       & 0.15 $\pm$ 0.09                      & 0.14 $\pm$ 0.06                       \\ \cline{1-1}
\multicolumn{1}{|l|}{Overall $l_1$} & \textbf{0.13 $\pm$ 0.07}             & 0.24 $\pm$0.12                        & 0.17$\pm$0.12                        & 0.25 $\pm$ 0.15                       & 0.18 $\pm$ 0.11                      & 0.25 $\pm$ 0.15                       & 0.16 $\pm$ 0.07                      & \textbf{0.21 $\pm$ 0.13}              \\ \cline{1-1}
\multicolumn{1}{|l|}{Overall $l_2$} & 0.16 $\pm$ 0.08                      & \textbf{0.15 $\pm$ 0.03}              & 0.14$\pm$0.06                        & 0.19 $\pm$ 0.08                       & \textbf{0.11 $\pm$ 0.03}             & 0.22 $\pm$ 0.05                       & 0.18 $\pm$ 0.09                      & 0.21 $\pm$ 0.08                       \\ \cline{1-1}
\multicolumn{1}{|l|}{Overall}       & 0.15 $\pm$ 0.06             & \textbf{0.19 $\pm$ 0.07}              & 0.155$\pm$0.08                       & 0.22 $\pm$ 0.1                        & \textbf{0.15 $\pm$ 0.05}             & 0.24 $\pm$ 0.08                       & 0.17 $\pm$ 0.07                      & 0.21 $\pm$ 0.11                       \\ \hline
\end{tabular}%
}
\end{table*}

\begin{table*}[!tbh]
\centering
\caption{Tabulated result of the individual parameter estimation error for \textit{articulated} objects with 2 links when the variance over the parameters converged after multiple interactions. The exploratory action is applied to $l_1$. For each interaction \textit{pulling} and \textit{pushing}, the best performing approach is highlighted in bold face.}

\label{tab:inference_articulated}
\resizebox{\textwidth}{!}{%
\begin{tabular}{l|ll|ll|ll|ll|}
                                    & \multicolumn{2}{c|}{A-GNN}                                                   & \multicolumn{2}{c|}{U-GNN}                                                   & \multicolumn{2}{c|}{R-GNN}                                                   & \multicolumn{2}{c|}{A-FF}                                                    \\ \cline{2-9} 
                                    & \multicolumn{1}{c}{\textit{Pushing}} & \multicolumn{1}{c|}{\textit{Pulling}} & \multicolumn{1}{c}{\textit{Pushing}} & \multicolumn{1}{c|}{\textit{Pulling}} & \multicolumn{1}{c}{\textit{Pushing}} & \multicolumn{1}{c|}{\textit{Pulling}} & \multicolumn{1}{c}{\textit{Pushing}} & \multicolumn{1}{c|}{\textit{Pulling}} \\ \cline{1-1}
\multicolumn{1}{|l|}{$m_{l_1}$}     & 0.08$\pm$0.09               & 0.15$\pm$0.13                         & 0.1$\pm$0.08                         & 0.32$\pm$0.22                         & 0.05$\pm$0.03                        & \textbf{0.13$\pm$0.14}                & \textbf{0.04$\pm$0.03}               & 0.16$\pm$0.12                \\ \cline{1-1}
\multicolumn{1}{|l|}{$m_{l_2}$}     & 0.23$\pm$0.11               & 0.22$\pm$0.1                          & 0.3$\pm$0.17                         & 0.25$\pm$0.13                         & \textbf{0.23$\pm$0.04}               & \textbf{0.18$\pm$0.06}                & 0.3$\pm$0.13                         & 0.22$\pm$0.11                \\ \cline{1-1}
\multicolumn{1}{|l|}{$f_{l_1}$}     & \textbf{0.09$\pm$0.01}               & 0.28$\pm$0.06                         & 0.1$\pm$0.02                         & \textbf{0.25$\pm$0.06}                & 0.11$\pm$0.02                        & 0.25$\pm$0.06                         & 0.1$\pm$0.01                         & 0.27$\pm$0.06                         \\ \cline{1-1}
\multicolumn{1}{|l|}{$f_{l_2}$}     & 0.18$\pm$0.01                        & 0.25$\pm$0.02                         & 0.19$\pm$0.07                        & \textbf{0.24$\pm$0.03}                & \textbf{0.16$\pm$0.01}               & 0.25$\pm$0.03                         & 0.17$\pm$0.04                        & 0.27$\pm$0.04                         \\ \cline{1-1}
\multicolumn{1}{|l|}{$CoM_{l_1}$}   & 0.07$\pm$0.05                        & \textbf{0.1$\pm$0.09}                 & 0.08$\pm$0.05                        & 0.14$\pm$0.06                & 0.11$\pm$0.04                        & 0.15$\pm$0.08                         & \textbf{0.05$\pm$0.02}               & 0.17$\pm$0.15                         \\ \cline{1-1}
\multicolumn{1}{|l|}{$CoM_{l_2}$}   & 0.15$\pm$0.09                        & 0.1$\pm$0.03                          & 0.17$\pm$0.09               & 0.12$\pm$0.09                & \textbf{0.13$\pm$0.04}               & 0.11$\pm$0.08                         & 0.16$\pm$0.13                        & \textbf{0.08$\pm$0.04}                \\ \cline{1-1}
\multicolumn{1}{|l|}{$f_j$}         & \textbf{0.11$\pm$0.11}               & \textbf{0.15$\pm$0.08}                & 0.16$\pm$0.11                        & 0.16$\pm$0.1                          & 0.13$\pm$0.1                         & 0.15$\pm$0.1                          & 0.12$\pm$0.11                        & 0.15$\pm$0.15                         \\ \cline{1-1}
\multicolumn{1}{|l|}{Overall $l_1$} & 0.12$\pm$0.05               & \textbf{0.23$\pm$0.05}                & 0.15$\pm$0.04                        & 0.29$\pm$0.08                         & 0.13$\pm$0.05                        & 0.23$\pm$0.07                         & \textbf{0.1$\pm$0.04}                & 0.25$\pm$0.06                \\ \cline{1-1}
\multicolumn{1}{|l|}{Overall $l_2$} & 0.19$\pm$0.06                        & 0.19$\pm$0.04                & 0.22$\pm$0.11                        & 0.2$\pm$0.05                          & \textbf{0.17$\pm$0.01}               & \textbf{0.18$\pm$0.03}                & 0.21$\pm$0.1                         & 0.19$\pm$0.05                         \\ \cline{1-1}
\multicolumn{1}{|l|}{Overall}       & 0.15$\pm$0.03               & \textbf{0.21$\pm$0.02}                & 0.18$\pm$0.07                        & 0.25$\pm$0.02                         & \textbf{0.15$\pm$0.02}               & 0.21$\pm$0.03                         & 0.16$\pm$0.05                        & 0.22$\pm$0.04                         \\ \hline
\end{tabular}%
}
\end{table*}
We also present the parameter estimation results of the daily object using our proposed approach $A-GNN$ and report the actual and predicted values, along with the predicted variance. 

\begin{table}[!tbh]
\caption{Tabulated result of comparison of the estimated parameters using $A-GNN$ along with the predicted variance from the filter for the daily object set.}
\label{tab:homo_daily}
\resizebox{\columnwidth}{!}{%
\begin{tabular}{l|ll|ll|ll|}
                                          & \multicolumn{2}{c|}{$m$ (kg)}                                         & \multicolumn{2}{c|}{$f$}                                              & \multicolumn{2}{c|}{$CoM$ (m)}                                        \\ \cline{2-7} 
                                          & \multicolumn{1}{c|}{\textbf{$GT$}} & \multicolumn{1}{c|}{\textbf{Pred}} & \multicolumn{1}{c|}{\textbf{$GT$}} & \multicolumn{1}{c|}{\textbf{Pred}} & \multicolumn{1}{c|}{\textbf{$GT$}} & \multicolumn{1}{c|}{\textbf{Pred}} \\ \hline
\multicolumn{1}{|l|}{\textit{sugar}}      & \multicolumn{1}{l|}{1.0}         & 1.12 $\pm$ 0.02                    & \multicolumn{1}{l|}{0.5}         & 0.12 $\pm$ 0.1                     & \multicolumn{1}{l|}{0.0}         & 0.0 $\pm$ 0.01                     \\ \cline{1-1}
\multicolumn{1}{|l|}{\textit{Nutella}}    & \multicolumn{1}{l|}{1.1}         & 1.2 $\pm$ 0.1                      & \multicolumn{1}{l|}{0.17}        & 0.12 $\pm$ 0.08                    & \multicolumn{1}{l|}{0.0}         & 0.01 $\pm$ 0.01                    \\ \cline{1-1}
\multicolumn{1}{|l|}{\textit{Paint}}      & \multicolumn{1}{l|}{1.61}        & 1.5 $\pm$ 0.05                     & \multicolumn{1}{l|}{0.3}         & 0.32 $\pm$ 0.05                    & \multicolumn{1}{l|}{0.0}         & 0.0 $\pm$ 0.01                     \\ \cline{1-1}
\multicolumn{1}{|l|}{\textit{Olive Oil}}  & \multicolumn{1}{l|}{1.24}        & 0.9 $\pm$ 0.02                     & \multicolumn{1}{l|}{0.15}        & 0.5 $\pm$ 0.12                     & \multicolumn{1}{l|}{0.02}        & 0.03 $\pm$ 0.01                    \\ \cline{1-1}
\multicolumn{1}{|l|}{\textit{Coffee}}     & \multicolumn{1}{l|}{0.67}        & 1.0  $\pm$ 0.06                    & \multicolumn{1}{l|}{0.2}         & 0.35 $\pm$ 0.1                     & \multicolumn{1}{l|}{0.0}         & 0.02 $\pm$ 0.03                    \\ \cline{1-1}
\multicolumn{1}{|l|}{\textit{Tuna}}       & \multicolumn{1}{l|}{0.35}        & 0.5 $\pm$ 0.08                     & \multicolumn{1}{l|}{0.25}        & 0.5 $\pm$ 0.2                      & \multicolumn{1}{l|}{0.0}         & 0.0 $\pm$ 0.02                     \\ \cline{1-1}
\multicolumn{1}{|l|}{\textit{Weight Box}} & \multicolumn{1}{l|}{0.82}        & 0.56 $\pm$ 0.1                     & \multicolumn{1}{l|}{0.3}         & 0.35 $\pm$ 0.1                     & \multicolumn{1}{l|}{0.03}        & 0.01 $\pm$ 0.03                    \\ \cline{1-1}
\multicolumn{1}{|l|}{\textit{Hammer}}     & \multicolumn{1}{l|}{1.2}         & 0.7 $\pm$ 0.25                     & \multicolumn{1}{l|}{0.2}         & 0.5 $\pm$ 0.1                      & \multicolumn{1}{l|}{0.07}        & 0.05 $\pm$ 0.08                    \\ \hline
\end{tabular}%
}
\end{table}

\subsubsection{Pose Tracking}
\begin{figure}[!tbh]
    \centering
    \includegraphics[width = 0.75\columnwidth]{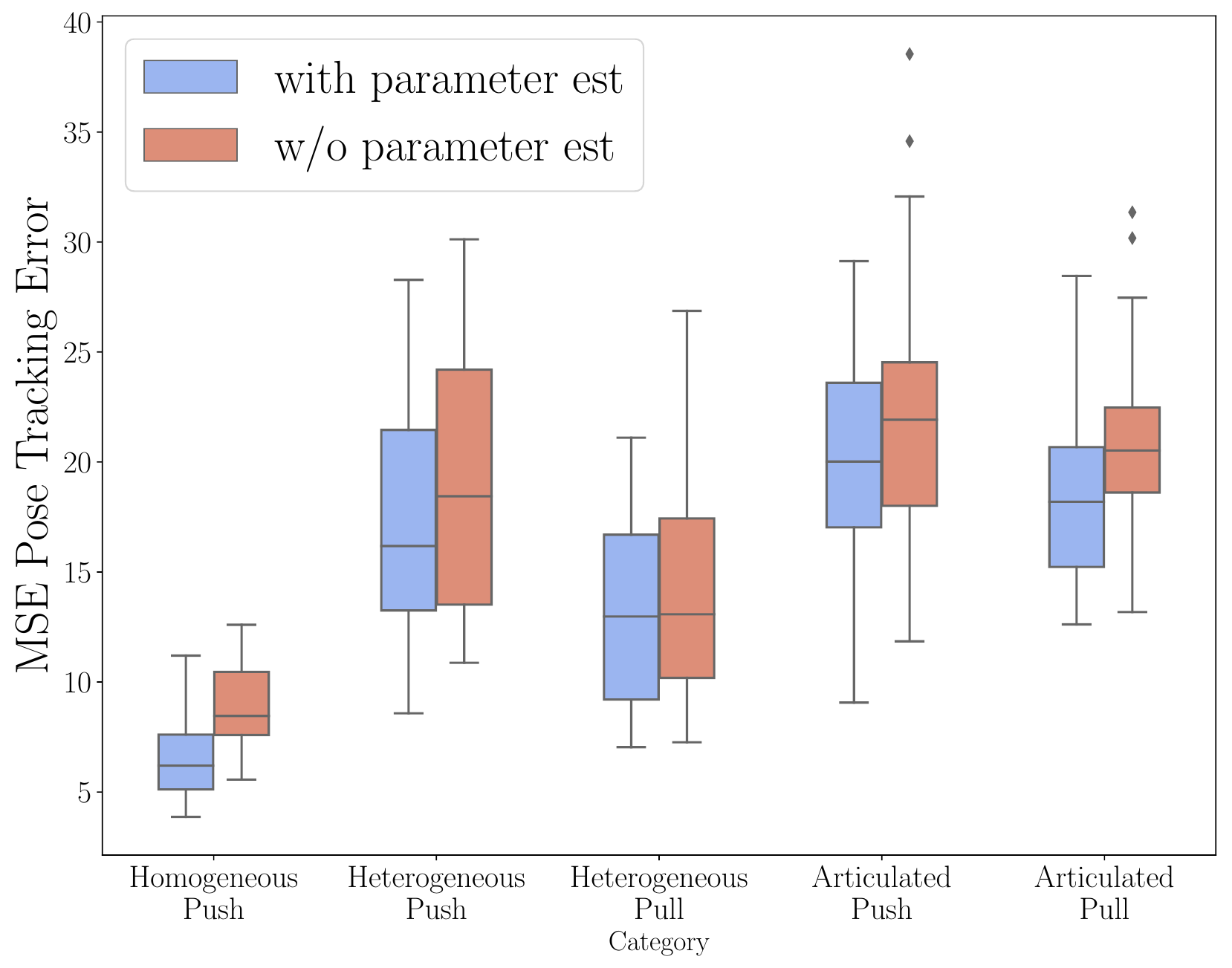}
    \caption{Pose tracking result of the proposed approach with and without parameter estimation for various objects and interactions.}
    \label{fig:pose_tracking_plot}
\end{figure}

As a critical application of parameter estimation, we present how pose estimation improves with and without parameter estimation for \textit{homogeneous}, \textit{heterogeneous}, and \textit{articulated} objects. We present a thorough result of all the test case objects undergoing different sampled pushing and pulling interactions. The bar graph in Fig. \ref{fig:pose_tracking_plot} shows the mean squared tracking error $MSE$, that is, the error w.r.t. $GT$ poses over the complete trajectory, along with the standard deviation computed over multiple exploratory trajectories/interaction. For the case of pose tracking without the estimated parameters, an initial estimate of the parameters was provided with low covariance to ensure that the dual differentiable filter performed only pose estimation/tracking. We can observe that the estimation of parameters significantly improves pose tracking across the different object types. The error in the \textit{pulling} interaction is lower than that of \textit{ pulling} due to the constrained movement of the object during the prehensile pulling. 

\subsubsection{Goal Driven Control}
\begin{figure}[t!]
    \centering
    \includegraphics[width = \columnwidth]{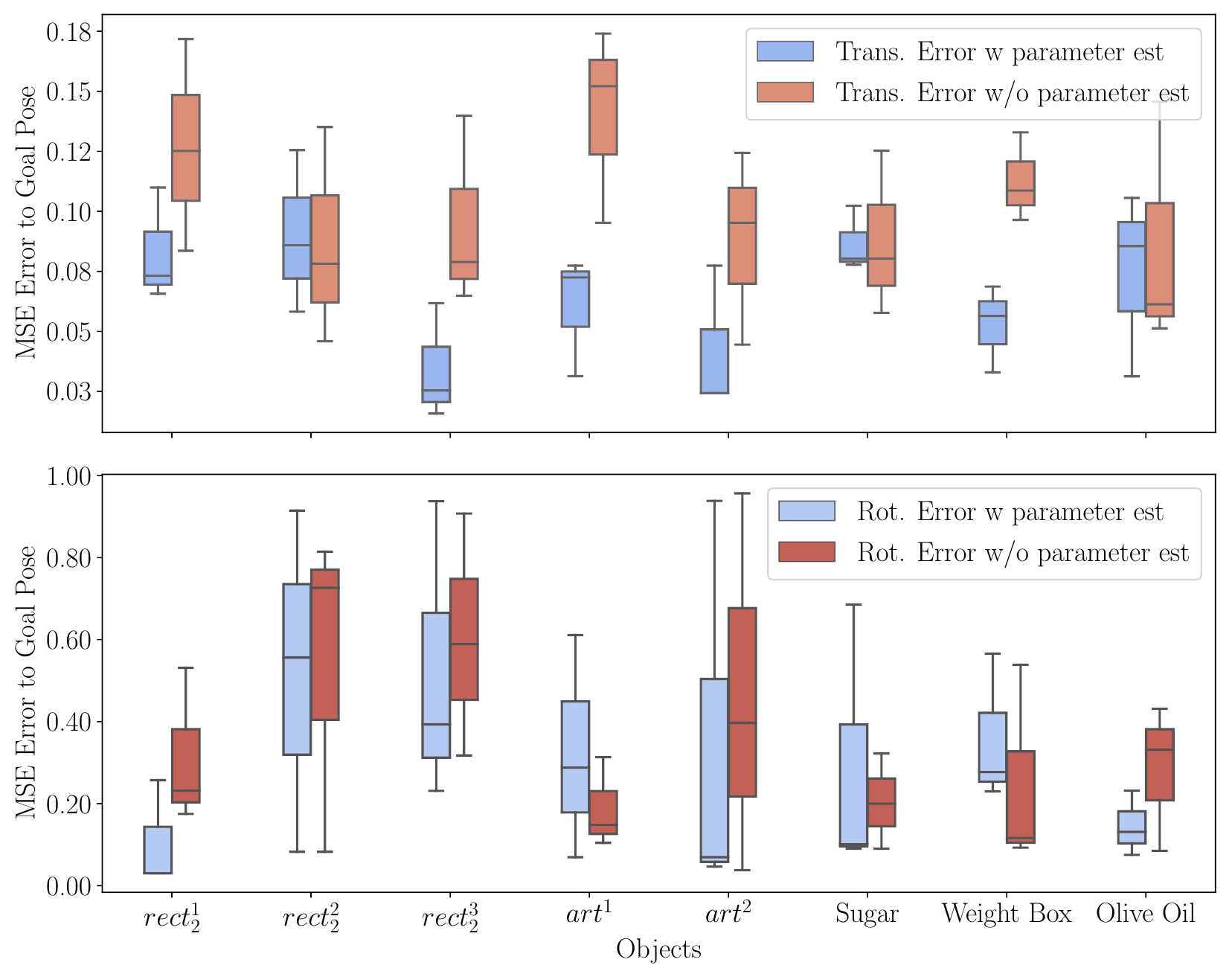}
    \caption{MSE of goal-driven control for 3 trials on 8 objects performed each i) with estimated parameters and ii) without the estimated parameters }
    \label{fig:control_plot}
\end{figure}

In this study, we used the learned process model to implement the iterative cross-entropy model predictive control (iCEM) approach \cite{pinneri2021sample} to showcase the effectiveness of parameter prediction in goal-driven control using non-prehensile push. Goal-driven pushing is a challenging task due to the complex interaction and unconstrained nature of the object, as compared to pulling an object towards a desired location. It is important to note that our main focus in this study is not on developing an optimal pushing control approach, and therefore, we did not compare it against state-of-the-art pushing control schemes. The iCEM approach utilizes the process model of the dual differentiable filter process model, which makes it suitable for our purposes. For the control experiment, we selected 8 objects and placed them in various initial locations, to push them towards a specific location $Goal Pose - [0.5, 0.1, 0]$ at the support location. The control scheme was executed 3 times for each object, once with estimated parameters and once without estimated parameters. Fig. \ref{fig:control_plot} illustrates a bar plot depicting the mean squared error (MSE) of the final goal pose. The results show that the estimation of physical parameters is crucial for downstream control tasks.  

\subsubsection{Change in Environment prediction}
Identifying changes or drifts in the environment from which data-driven models were trained is a crucial concern. We attempt to address this challenge and demonstrate that our proposed predictive framework has the potential to address this issue. From eq. \ref{eq:parameter_update} we can compute the likelihood of observation as follows. 

\begin{flalign}
    ObsLike &=  e^{(-\frac{1}{2}(\overline{\mathcal{Z}}^{[j]}_t - z_t) R^{-1}(\overline{\mathcal{Z}}^{[j]}_t - z_t)^T)}
     \label{eq:obs_like}
\end{flalign}

\begin{figure}[t!]
    \centering
    \includegraphics[width = 0.75\columnwidth, height=6cm]{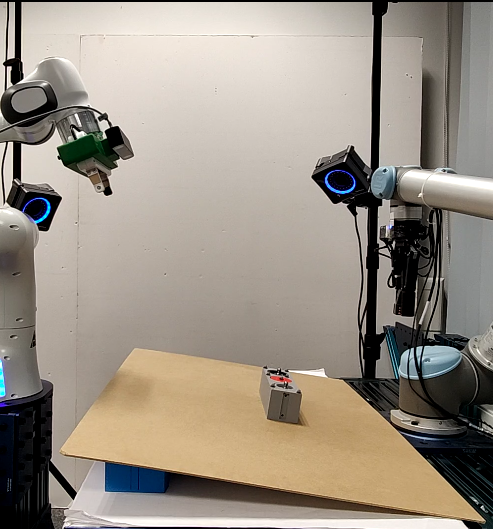}
    \caption{Setup for inducing a change in the learned environment by tilting the support surface.}
    \label{fig:change_env}
\end{figure}

Using the above equation, when the parameters are estimated and the object-robot is interacting in the same environment as it was trained in, the likelihood of new observations is expected to be high. However, whenever there is a change in the environment, the new observations are likely to shift, resulting in a lower likelihood prediction. Previous approaches have not been able to utilize this likelihood of observation, as the learned model was quite inaccurate, the noise was not captured effectively, or the parameters were not incorporated correctly. In this work, we attempted to elevate these issues and explored whether the observation likelihood can be a valuable indicator for detecting the change in the environment, much like how humans predict changes. In all our training setups, the support was kept flat; however, we purposely changed the support angle as shown in Fig. \ref{fig:change_env}.

\begin{figure}[t!]
    \centering
    \includegraphics[width = \columnwidth]{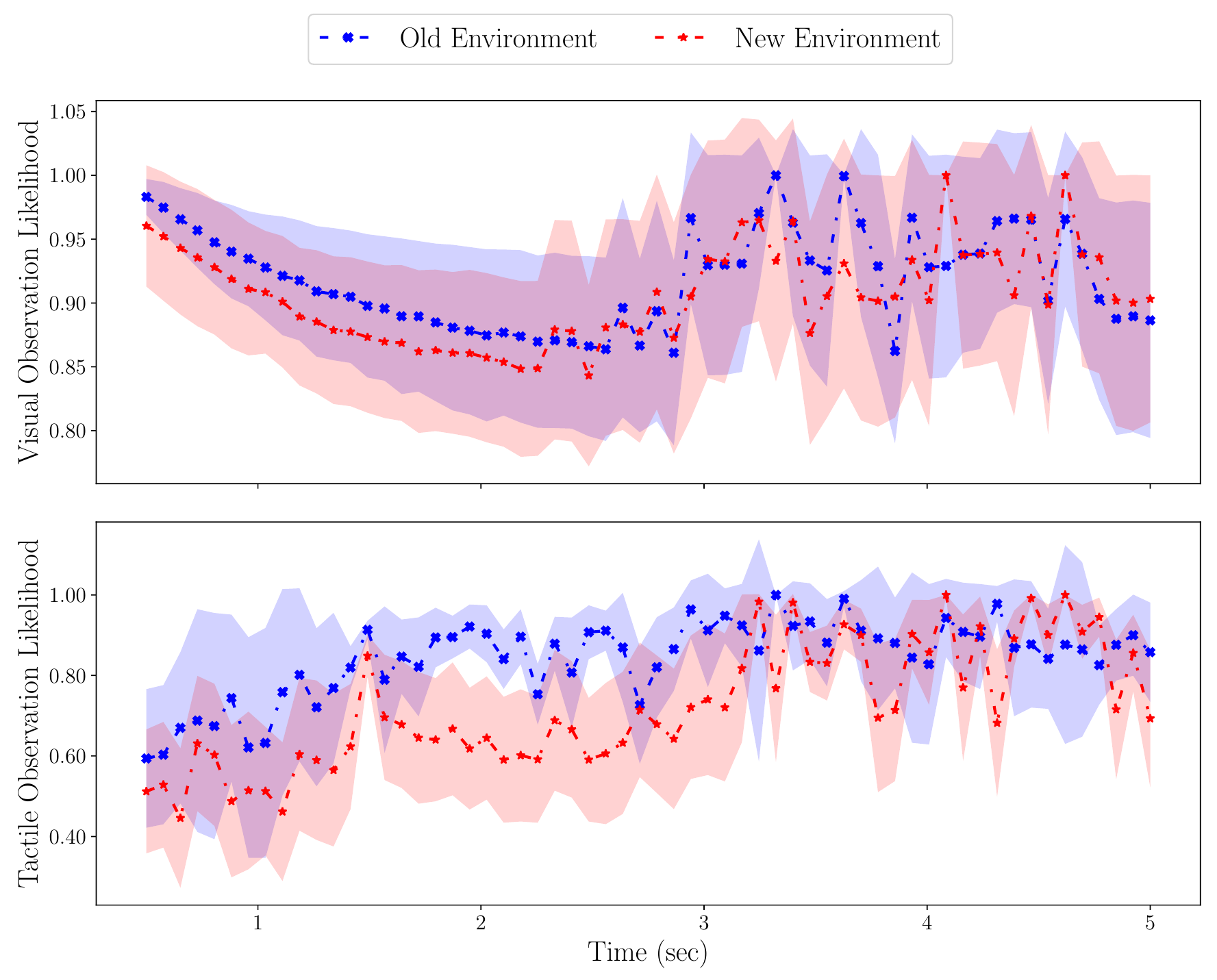}
    \caption{Likelihood of visual and tactile observation comparing flat surface interaction (old environment) with tilted surface interaction (new environment)}
    \label{fig:ood_prediction}
\end{figure}

We executed multiple exploratory push trajectories in this setup for a known (parameters are estimated) \textit{homogeneous} object in a tilted and flat surface and computed the likelihood of visual and tactile observation using Eq.\ref{eq:obs_like}. Figure. \ref{fig:ood_prediction}  presents the computed likelihood along with the standard deviation. It can be observed that tactile observation likelihood can reliably detect the change in the environment, as we observe a decrease in likelihood from 2-3 seconds. This creates an opportunity to retrain the interaction model when the system identifies significant changes in the environment.

%% file: sections/conclusion.tex
\subsection{Discussion}
\label{subsec:discussion}
In this work, we proposed a novel interactive visuo-tactile framework for inferring the physical properties of diverse unknown rigid objects without prior information. Through our framework robotic system uses visual information to estimate object shapes via superquadrics and Bayesian inference, determining the next-best view for complete shape information in complex scenarios. This approach avoids occupancy grid approximations and accurately calculates entropy, considering the intrinsic settings of the camera. After shape and pose estimation, the robotic system uses visual and tactile sensing to estimate inertial and frictional properties through prehensile pulling or non-prehensile pushing. Our dual differentiable filtering approach estimates both time-invariant object parameters and time-variant object pose. The innovative $N$-step look-ahead formulation optimizes exploration actions by considering object state uncertainty. We also introduced a novel graphical representation of the interaction between objects and robots using a graph neural network (GNN) within the dual-differentiable filter, to autonomously handle the diverse types of unknown objects - \textit{homogeneous, heterogeneous, articulated}

To evaluate the performance of our proposed framework, we designed a set of diverse objects with \textit{homogeneous}, \textit{heterogeneous} and \textit{articulated} objects. We had 120 configurable 3D printed objects with diverse shapes, inertial, and frictional properties as presented in Fig. \ref{fig:experimental_object_set}. In addition, we selected $8$ daily objects with varying properties as shown in Fig. \ref{fig:experimental_object_set}.


The experimental results (see Fig.\ref{fig:shapeplot}) demonstrate that our proposed active shape perception approach could robustly estimate the shapes of objects (see Fig.\ref{fig:experimental_object_set}) with Chamfer distance ($CD$) $< 0.002$. Moreover, Fig.\ref{fig:activeshape} shows that the \textit{active} approach can estimate the superquadric parameters in $3-4$ viewpoints with low-variance, compared to baseline \textit{uniform} or \textit{random} viewpoint selection approach. In addition, Fig. \ref{fig:qualshapeplot} illustrates the qualitative results of the estimated superquadrics after using active visual perception, which proved to be robust for interactive exploration. Also, the proposed nonlinear deformation factors in superquadrics adequately capture complex shapes such as $tri_1$, $butter_1$, and $butter_2$. In cases of \textit{heterogeneous} objects have a lower shape accuracy (high $CD$), for example, $rect_1-rect_3$ has $CD$ error of $0.002$ compared when $rect_3$ as homogeneous or $ellip_2-rect_3$ as articulated object. This is expected due to the close vicinity of the combined noisy point clouds obtained for \textit{heterogeneous} objects. 

To infer inertial and frictional properties with the proposed dual differentiable filtering approach, iterative training of the graph neural network (GNN) model and observation noise networks was performed. Fig. \ref{fig:learningplot} shows that the proposed GNN model coupled with active action selection has a significant improvement in data efficiency; this is especially evident in complicated articulated object and push interaction, where the proposed method had an efficiency of $25\%$ over uniform action selection and $9\%$ over the baseline $A-FF$ with active action and feedforward model. 

Fig.\ref{fig:learned_noise_models} illustrates the effectiveness of the learned model in capturing the heteroscedasticity of the visual and tactile noise variance for both prehensile and non-prehensile interaction in the setup. The results in Fig.\ref{fig:learned_noise_models} show that tactile noise is much lower than visual noise, especially in the case of prehensile interaction. This also enables the filter to appropriately handle visual and tactile observations for estimation. The results achieved (see Figs. \ref{fig:inferencehomo}, \ref{fig:inference_hetero}, \ref{fig:inference_articulated})  for different object parameter estimation with specific interactions highlight that the proposed active action affordance selection with the graph neural network $A-GNN$ outperforms (lower $NRMSE$) baseline $A-FF$. Although for some parameters, A-FF performs better than A-GNN due to inherently different network structure, however, the proposed GNN based approach elevates the limitations of requiring the number of links of objects to be known beforehand.

The outcome of the ablation study with uniform $U-GNN$ and random $R-GNN$ action affordances shows that the active approach performed consistently across different types of interaction and various objects. The role of active action in the case of the prehensile interaction was found to be more significant than that of the pushing interaction (see Table.\ref{tab:inference_hetero}  and \ref{tab:inference_articulated}). This could be because the dynamics of the prehensile interaction are more constrained compared to the non-prehensile, leading to lower excitation of the object state \cite{mason_pushing}, which can also be seen by slightly higher errors in the prehensile interaction compared to the non-prehensile interaction. However, it is desirable to have the option of having both interactions to exploit the constraints of the robotic workspace. 

\textit{Articulated} objects compared to \textit{homogeneous} or \textit{heterogeneous} has higher error rates of parameter estimation. It also requires a higher number of interactions during the training phase. These are due to the complexity of the rotational degree of freedom between the object links. In addition, the estimation of parameters of the object link which was not in contact with the robot was challenging as there was limited tactile information transferred through the joint and relied solely on visual information. It can also be observed that the error variance of parameter estimation of \textit{heterogeneous} objects was high, due to confounding parameters leading to similar visual and tactile observations. In addition, for \textit{homogeneous} objects with pushing interaction, the analytical approach $A-Analytical$ had much better accuracy for the center of mass $CoM$ estimation but with high errors for mass and friction. 

We also demonstrated the importance of the parameter estimation problem by demonstrating substantial improvement in pose estimation during the interactions (see Fig. \ref{fig:pose_tracking_plot} and also in goal-driven pushing (see Fig.\ref{fig:control_plot}). We also attempted to address the challenging problem of detecting a change in the learned model by exploiting the predictive and Bayesian formulation of our proposed approach. Fig.\ref{fig:ood_prediction} illustrates how tactile observations prove to be a strong indicator of detecting `out of distribution' object-robot interaction. 

The experiments in this study were conducted on a workstation running Ubuntu 18.04, equipped with an Intel Xeon(R) CPU 5222 @ 3.80 GHz, 32 GB RAM, and an NVIDIA Quadro RTX 4000. Due to the computational complexity (resource-hungry nature) of our proposed framework, which is based on sampling-based techniques, we were able to achieve a maximum execution time of $5 Hz$ for the active dual differentiable filter and $1 Hz$ for active shape perception. In future research, it would be valuable to investigate ways to further reduce the computational complexity, and improve the perception frequency.

In this study, our objective was to estimate the inertial and frictional properties of various rigid objects through push or pull actions. Future work will extend our method to objects with dynamic properties, such as changing centers of mass or deformable structures. Additionally, we aim to evaluate the use of simple and natural prehensile and non-prehensile interactions in a bimanual manner. Further research is needed in active inference, particularly in complex scenarios involving heterogeneous and articulated objects, where active action selection was less effective compared to homogeneous objects. Moreover, our current shape perception approach is not suitable for more complex objects with shallow interiors, such as cups and bowls, highlighting the need for a more robust and generalizable shape perception technique, which remains an active area of research.

Here, we consider the case where a single object is present for exploration; however, in the future, we will consider more challenging scenarios when the object is in clutter. We would like to integrate our previous decluttering approach as in \cite{vt_praj_2020} and take advantage of the decluttering interaction to estimate the properties of objects. 

\section{Conclusion}
In this paper, we address the problem of estimating the properties of various homogeneous, heterogeneous, and articulated rigid objects using visual and tactile sensing. The key innovations of our proposed interactive perception framework are active shape perception, active interaction action affordance selection, and dual differentiable filters with graphical models. Importantly, the framework allows the robotic system to estimate the properties of novel objects autonomously using simple interactive actions: non-prehensile push and prehensile pull. Our proposed approach was extensively validated on a comprehensive set of planar rigid objects on a real-robotic platform. The experiments demonstrated that the proposed approach outperforms the baseline approach and overcomes the limitations of previous studies from the literature. Furthermore, the efficacy of our framework was demonstrated in the three main applications of object tracking, goal-driven push, and detecting change in the environment. We believe that a learnable graphical model incorporated within the filtering formulation opens up the possibility of generalizing to different robotic setups and other estimation problems. In the future, we plan to examine the potential of our framework using a hierarchical interactive method to explore numerous objects and their corresponding environments employing shared visuo-tactile perception.



%% file: sections/appendix.tex
\section{Appendix}
\label{app:anal_model}
\subsection{Analytical Model of pushing}
Objects with \textit{homogeneous} properties have an analytical model of quasistatic pushing created by Lynch et al. \cite{lynch1992manipulation}. This model is based on several assumptions, but it helps to comprehend which physical parameters are involved in non-prehensile pushing. Here, we present the model for reference.

\subsubsection*{Analytical model of Non-prehensile pushing}
The analytical model of quasi-static pushing predicts the movement of the object $v_t$, given the velocity of the push of the robot $(u)$, the contact point $(cp)$ and the normal surface associated at the contact point $n$, as well as surface friction $\mu$ and friction between the object and the robot $\mu_r$. Predicting the effect of the push with this model has two stages. 

First, we determine whether the contact between the object and the robot is slipping or sticking, and then we compute the effective push velocity. 

\begin{figure}[!htb]
    \centering
    \includegraphics[width=0.6\columnwidth]{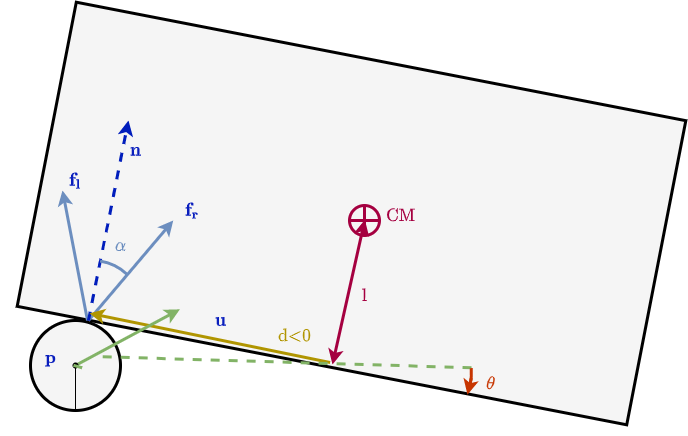}
    \caption{Analytical model of planar pushing}
    \label{fig:push_anal}
\end{figure}

\begin{align}
    & \alpha = arctan(\mu_r) \\
    & \mathbf{f_l} = \mathbf{R}(-\alpha)\mathbf{n} \\
    & \mathbf{f_r} = \mathbf{R}(\alpha)\mathbf{n} \\
    & m_l = cp_{x}f_{ly}-cp_{y}f_{lx} \\
    & m_r = c_{x}f_{ry}-c_{y}f_{rx}
\end{align}

The rotation matrix $\mathbf{R}(\alpha)$ represents a rotation about the $z$-axis with an angle $\alpha$. The contact point $cp$ and the surface normal $\mathbf{n}$ are relative to the center of mass of the object. Lynch et al. \cite{lynch1992manipulation} used an ellipsoidal approximation of the limit surface to relate the forces to the velocities of the object. To simplify the notation, the subscript $b$ is used to refer to either the left ($l$) or right ($r$) boundary forces. Linear and angular object velocities are denoted by $v_{o,b}$ and $\omega_{o,b}$, respectively. The push velocities that would be created from the boundary forces are referred to as $v_{p,b}$ and form the so-called "motion cone" as shown in Figure.\ref{fig:push_anal}.

\begin{align}
  & v_{o,b} = \frac{\omega_{o, b}l^2}{m_b}f_b \\
  & v_{p,b} = \omega_{o, b}(\frac{l^2}{m_b}f_b + k \times cp) 
\end{align}

Here, $k$ is the rotation of the object. where $\omega_{o,b}$ acts as a scaling factor. As we are only interested in the direction of $v_{p,b}$ and not in its magnitude, we set $\omega_{o,b}=m_b$:

\begin{equation}
     v_{p,b} = l^2f_b + m_b k \times c^{'}
\end{equation}

To compute the effective push velocity $v_p$, we need to determine the contact case: If the push velocity is outside the motion cone, the contact will slip. The effective resulting push then acts in the direction of the boundary velocity $v_{p,b}$, which is closer to the push direction:

\begin{equation}
     v_{p} = \frac{u n}{v_{p,b} n} v_{p,b}
\end{equation}

Otherwise, the contact sticks, and we can use the pusher velocity as the effective push velocity $v_p = u$. Now, given the effective push velocity in both the stick case and the slip case, we can compute the linear and angular velocity $v_o=[v_{ox}, v_{oy}, \omega]$

\begin{align}
    & v_{ox} = \frac{(l^{2} + c^{'2}_{x})v_{px} + c^{'}_{x} c^{'}_{y}}{l^{2} + c^{'2}_{x} + c^{'2}_{y}}v_{px} \\
    & v_{oy} = \frac{(l^{2} + c^{'2}_{x})v_{px} + c^{'}_{x} c^{'}_{y}}{l^{2} + c^{'2}_{x} + c^{'2}_{y}}v_{px} \\
    & \omega = \frac{c^{'}_{x} v_{oy} - c^{'}_{y} v_{ox}}{l^{2}}
\end{align}


\subsection{Qualitative shape perception results}
\label{app:qualshape}
\begin{figure}[!htb]
    \centering
    \includegraphics[width = 0.8\columnwidth]{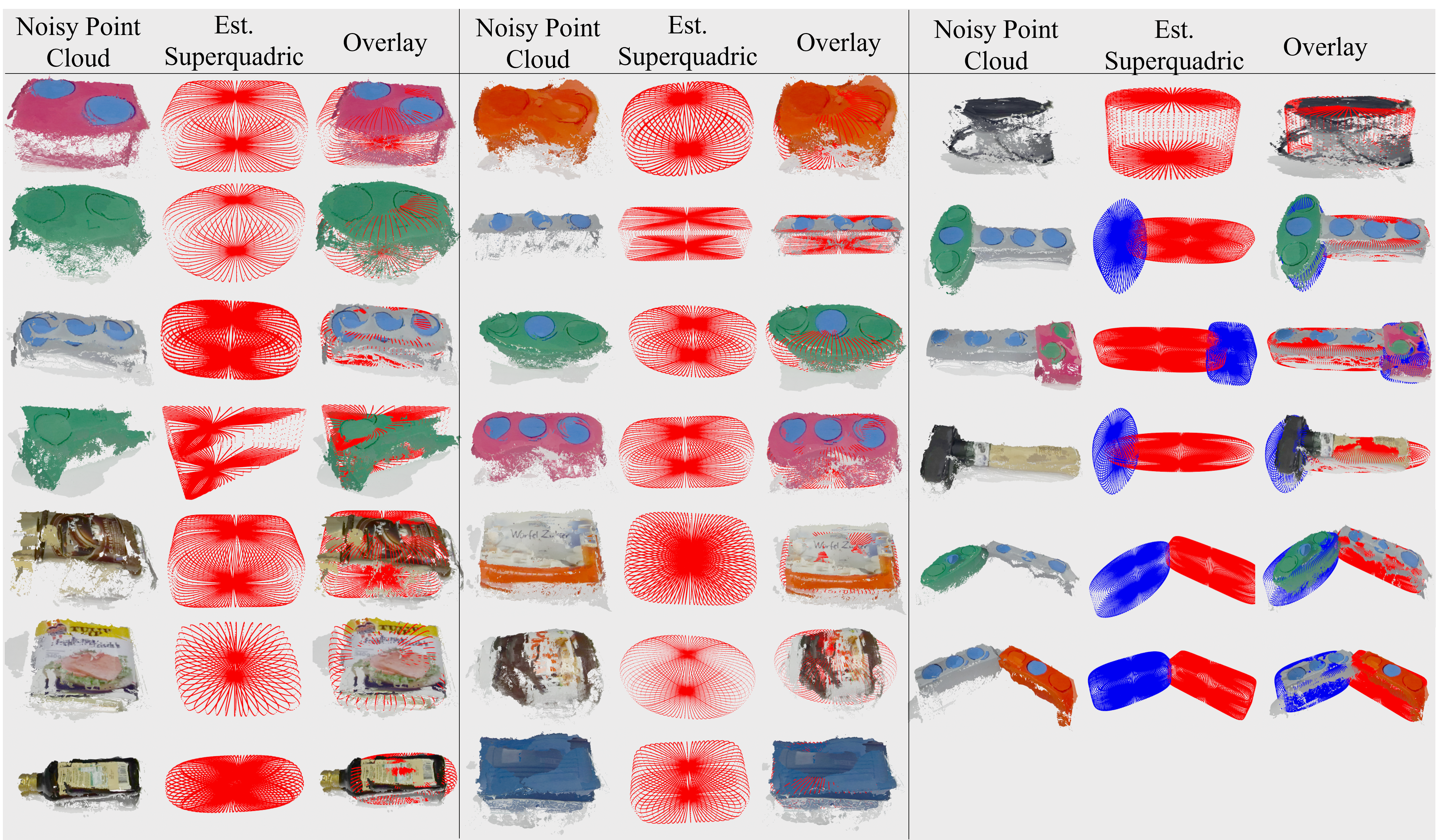}
    \caption{Qualitative shape perception result of the object set considered in this work}
    \label{fig:qualshapeplot}
\end{figure}

In this section we present the qualitative plot for all the shapes considered in the work. Fig.\ref{fig:qualshapeplot} presents the noisy and partial point cloud obtained after multiple views, the estimated superquadric shape, and the overlaid visualization for better illustration. We can observe the shape represents the actual object sufficiently for interactive object exploration. 

\begin{figure}[!tbh]
    \centering
    \includegraphics[width = \columnwidth]{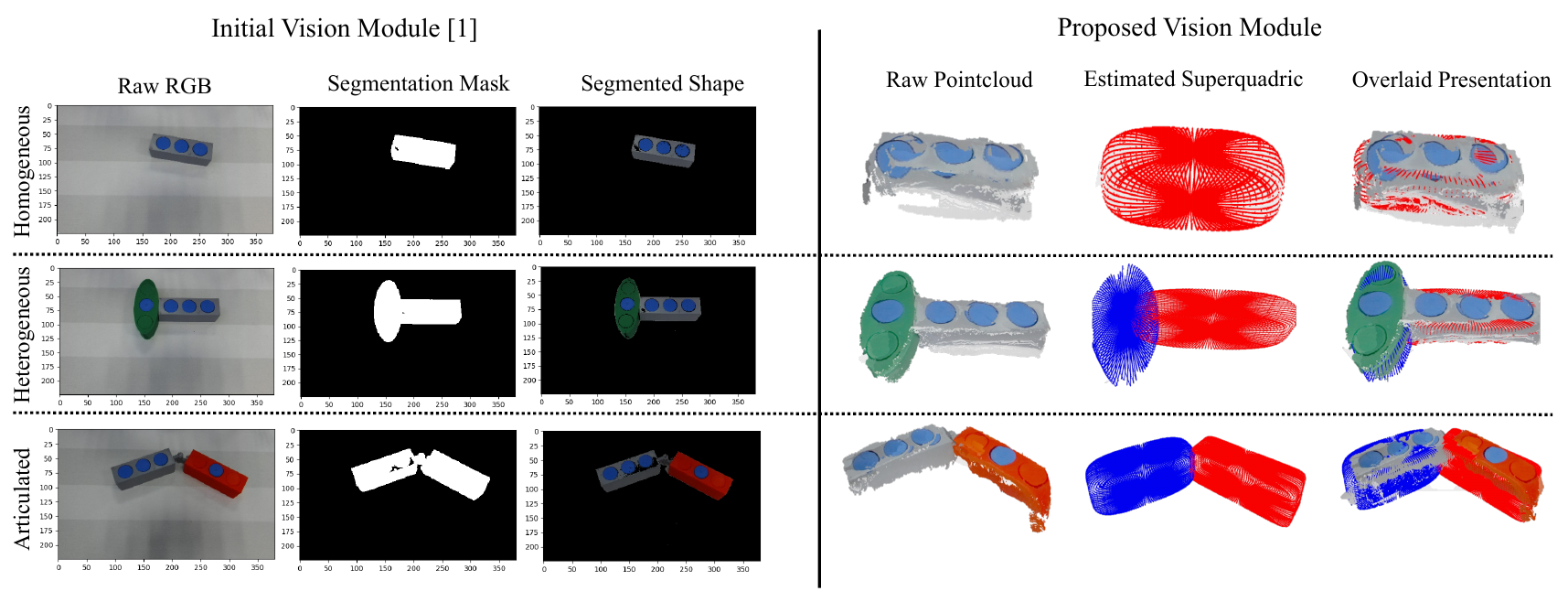}
    \caption{Comparision of the vision module of our previous work \cite{dutta2023push} and the newly proposed approach.}
    \label{fig:shape_comaprison}
\end{figure}

In addition, we present a experimental evaluation of the initial shape perception approach in our prior work in \cite{dutta2023push} to that of the proposed work in Fig.\ref{fig:shape_comaprison}

\subsection{Dual Differentiable Filtering}
\begin{figure}[!htb]
    \centering
    \includegraphics[width = 0.6\columnwidth]{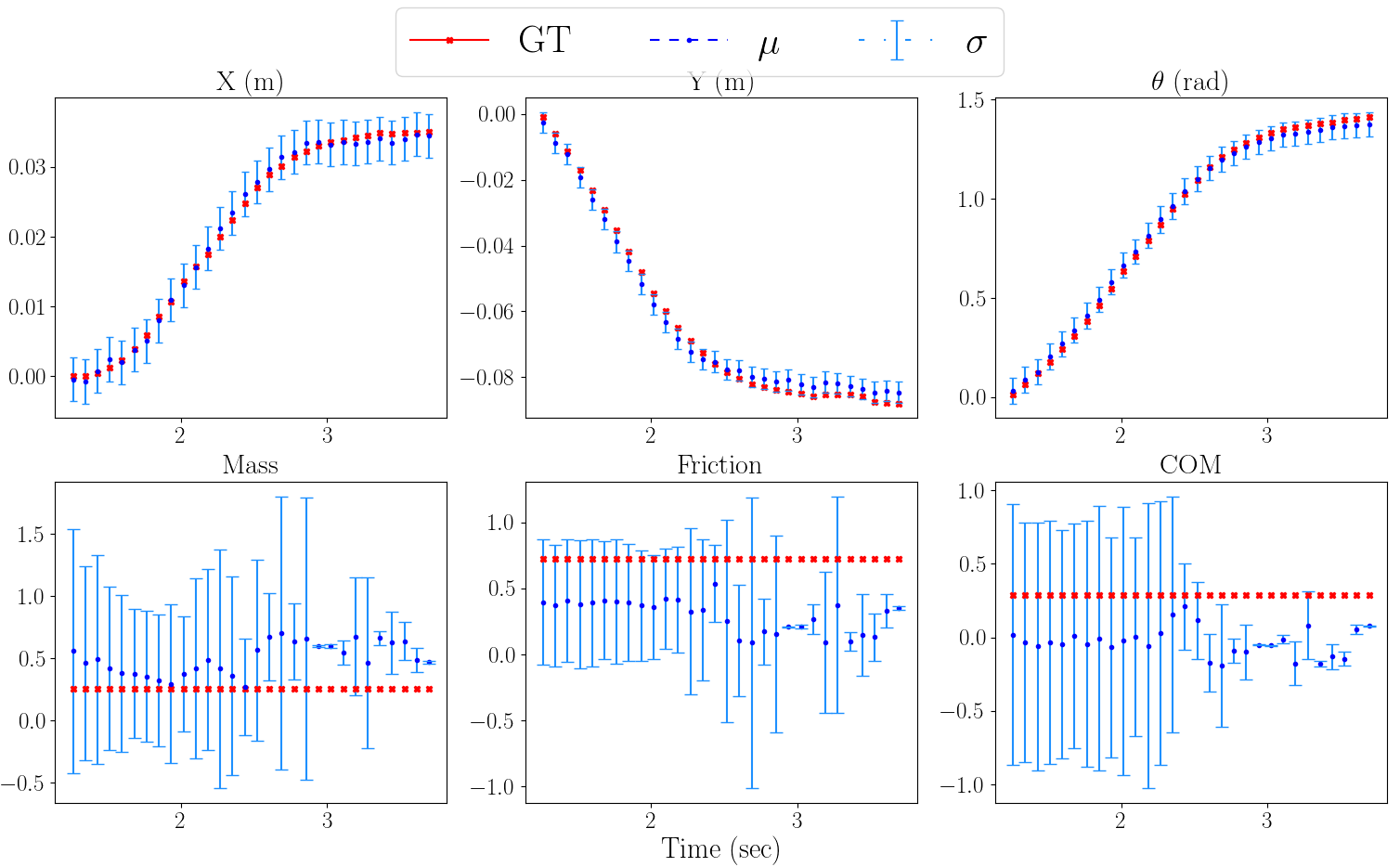}
    \caption{Examination of the filtering process of a representative trajectory for the case of \textit{homogeneous} object undergoing non-prehensile push interaction.}
    \label{fig:filteringuniform}
\end{figure}

\begin{figure}[!htb]
    \centering
    \includegraphics[width = 0.7\columnwidth]{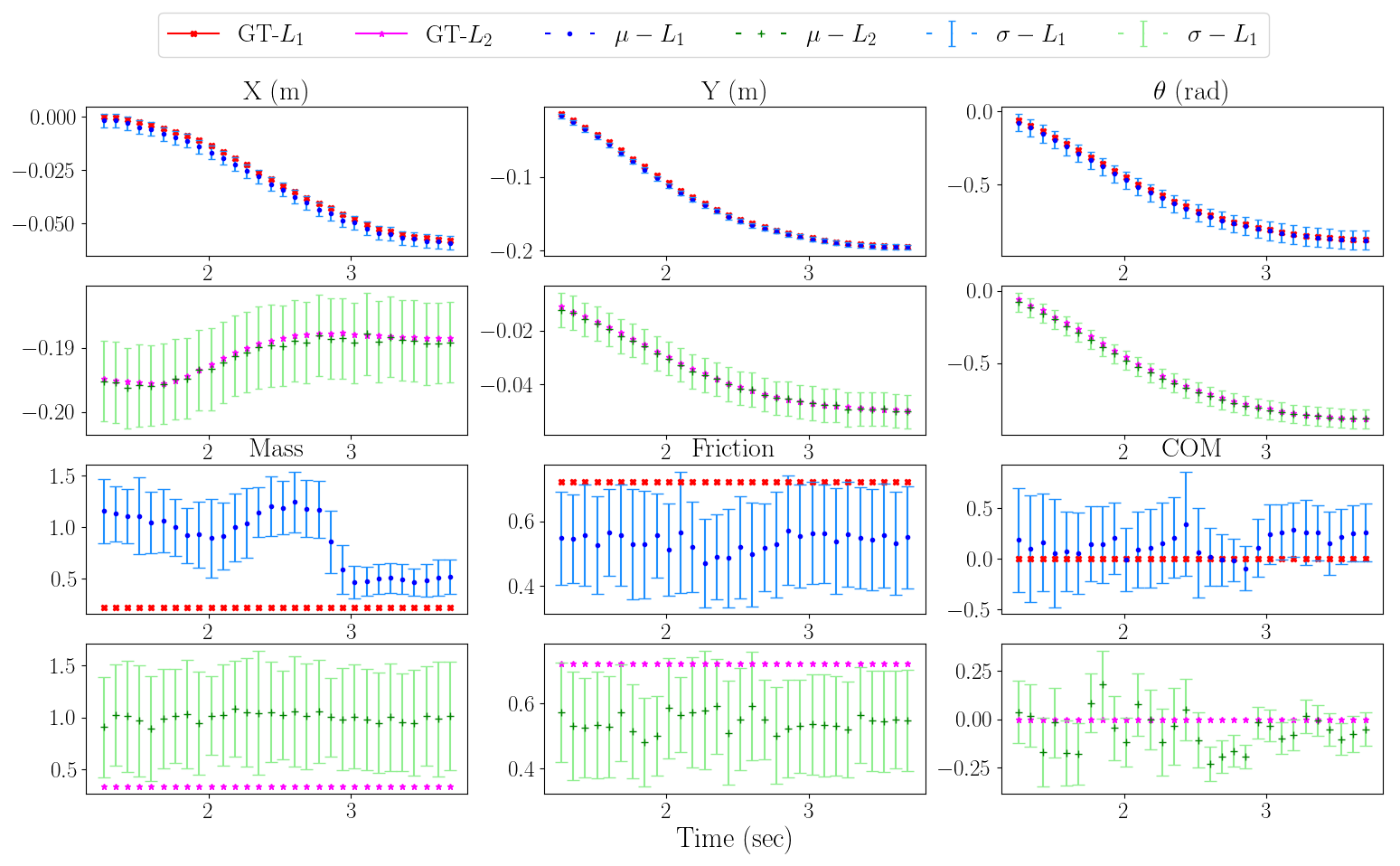}
    \caption{Examination of the filtering process of a representative trajectory for the case of \textit{heterogeneous} object with links $l_1$ and $l_2$, undergoing non-prehensile push interaction.}
    \label{fig:filteringheteropush}
\end{figure}

\begin{figure}[!htb]
    \centering
    \includegraphics[width = 0.7\columnwidth]{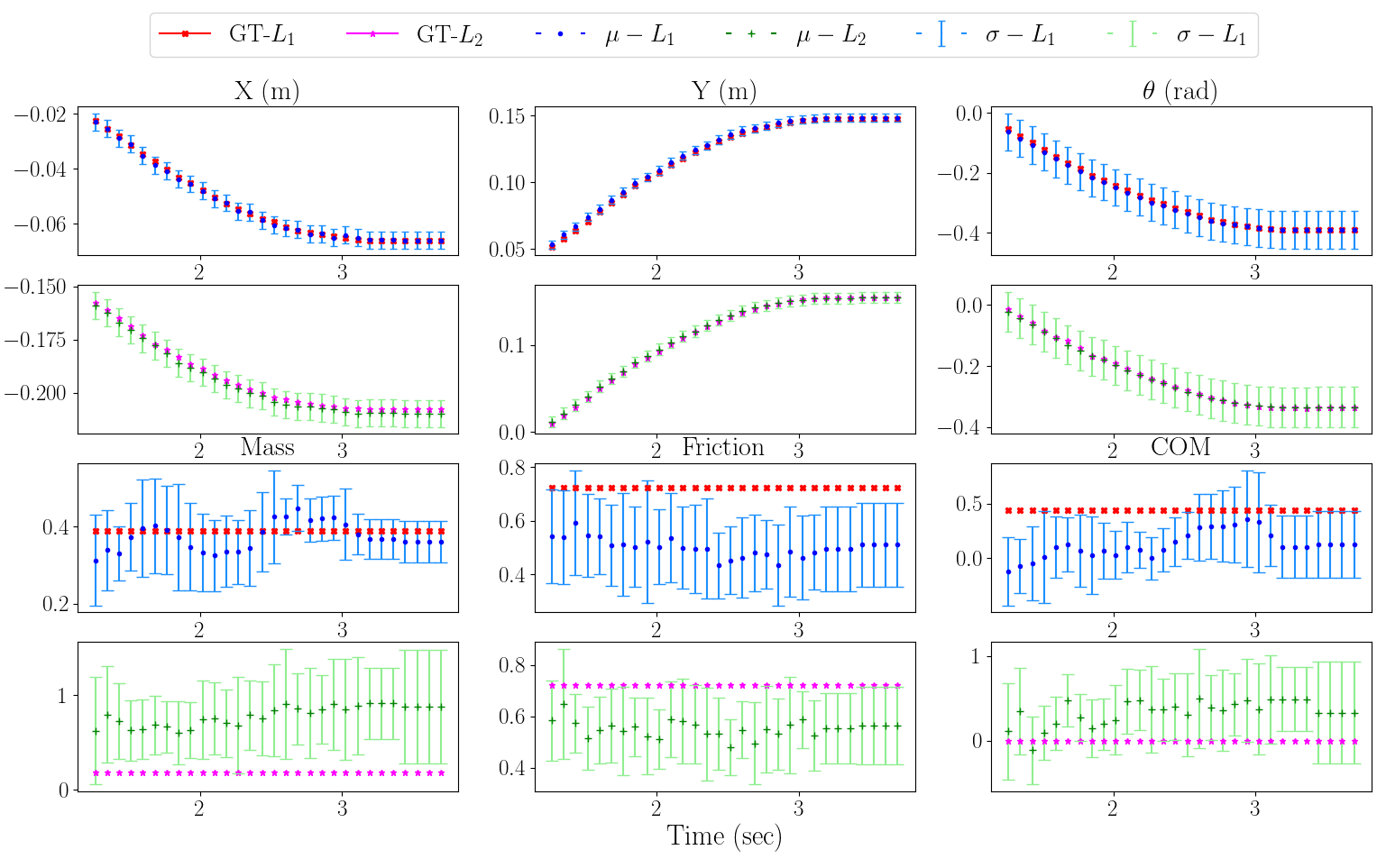}
    \caption{Examination of the filtering process of a representative trajectory for the case of \textit{heterogeneous} object with links $l_1$ and $l_2$, undergoing prehensile pull interaction.}
    \label{fig:filteringheteropull}
\end{figure}

\begin{figure}[!htb]
    \centering
    \includegraphics[width = 0.7\columnwidth]{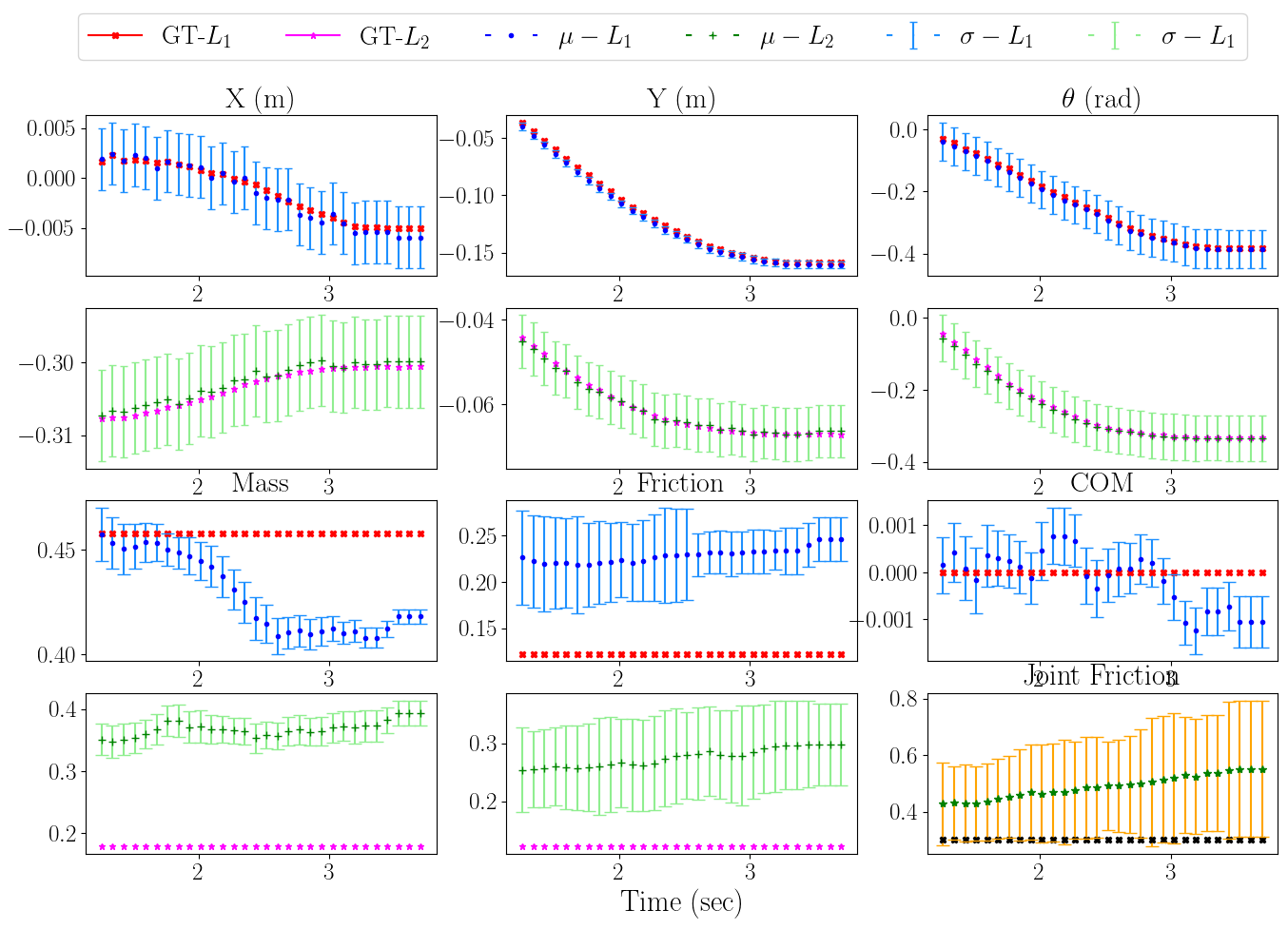}
    \caption{Examination of the filtering process of a representative trajectory for the case of \textit{articulated} object with links $l_1$ and $l_2$, undergoing non-prehensile push interaction.}
    \label{fig:filteringarticulatedpush}
\end{figure}

\begin{figure}[!htb]
    \centering
    \includegraphics[width = 0.7\columnwidth]{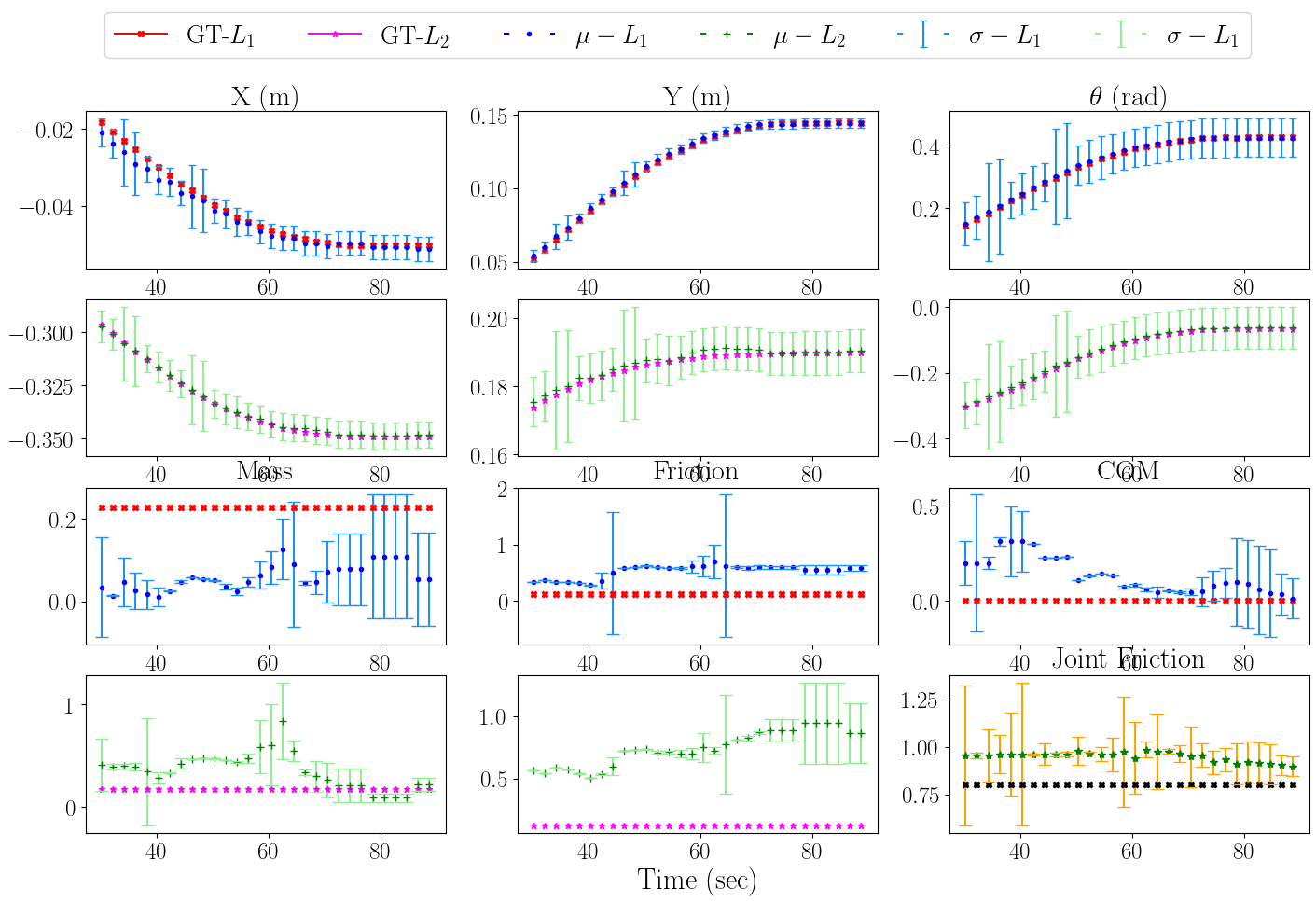}
    \caption{Examination of the filtering process of a representative trajectory for the case of \textit{articulated} object undergoing prehensile pull interaction.}
    \label{fig:filteringarticulatedpull}
\end{figure}

Fig. \ref{fig:filteringuniform} illustrates a detailed examination of the dual filtering process using the proposed $GNN$ approach. The plots show the filtering of pose and parameter state of a representative test case that involving a \textit{homogeneous} object with a non-prehensile pushing trajectory for exploration. Similarly, Figs. \ref{fig:filteringheteropush} and \ref{fig:filteringheteropull} depict the filtering procedure for a \textit{heterogeneous} object with non-prehensile pushing and prehensile pulling, respectively. Additionally, Figs. \ref{fig:filteringarticulatedpush} and \ref{fig:filteringarticulatedpull} present the filtering outcome for an \textit{articulated} object. These plots provide insights into the evolution of parameter variance over time and its impact on object pose estimation and filtering. The ground truth pose and the parameter are denoted as $GT$. This filtering process is carried out over multiple interactions until the variance over the parameters is reduced.

\vspace{-3pt}

